\documentclass{article}
\usepackage{amsmath,graphicx}
\usepackage{caption}
\usepackage{subcaption}
\usepackage{amssymb}
\usepackage{algorithm}
\usepackage{algorithmic}
\usepackage{color, textcomp}
\usepackage{booktabs, multirow}
\usepackage[utf8]{inputenc}

\newcommand{\R}{\ensuremath{\mathbb{R}}}
\newcommand{\V}{\ensuremath{\mathcal{V}}}
\newcommand{\E}{\ensuremath{\mathcal{E}}}
\newcommand{\G}{\ensuremath{\mathcal{G}}}
\newcommand{\N}{\ensuremath{N}}
\newcommand{\Ns}{\ensuremath{N_s}}
\newcommand{\Nt}{\ensuremath{N_t}}
\newcommand{\Ms}{\ensuremath{M_s}}
\newcommand{\Mt}{\ensuremath{M_t}}

\newcommand{\K}{\ensuremath{Q}}

\def\R{{\mathbb{R}}}

\title{Graph Domain Adaptation with Localized Graph Signal Representations}
\author{Yusuf Yi\u git Pilavc\i, Eylem Tu\u g\c ce G\"uneyi, Cemil Cengiz and Elif Vural
\thanks{Y. Y. Pilavc{\i} is with the GIPSA Lab at Universit\'e Grenoble Alpes, Grenoble. E. T. G\"uneyi and E. Vural are with the Dept.~of Electrical and Electronics Engineering at METU, Ankara. C. Cengiz is with the Dept.~of Computer Science and Engineering at Ko{\c c} University, Istanbul.
Most part of this work was performed while the authors were at METU.}}
\begin{document}
\date{}	
	
\maketitle

\abstract{In this paper we propose a domain adaptation algorithm designed for graph domains. Given a source graph with many labeled nodes and a target graph with few or no labeled nodes, we aim to estimate the target labels by making use of the similarity between the characteristics of the variation of the label functions on the two graphs. Our assumption about the source and the target domains is that the local behaviour of the label function, such as its spread and speed of variation on the graph, bears resemblance between the two graphs. We estimate the unknown target labels by solving an optimization problem where the label information is transferred from the source graph to the target graph based on the prior that the projections of the label functions onto localized graph bases be similar between the source and the target graphs. In order to efficiently capture the local variation of the label functions on the graphs, spectral graph wavelets are used as the graph bases. Experimentation on various data sets shows that the proposed method yields quite satisfactory classification accuracy compared to reference domain adaptation methods.

}

\section{Introduction}
	
A common assumption in  machine learning is that the training and the test data are sampled from the same distribution. However, in many practical scenarios the distributions of data samples in the training and test phases may differ. Domain adaptation methods aim to provide solutions to machine learning problems by dealing with this distribution discrepancy. In domain adaptation, the label information is mostly available in a source domain, while few or no class labels are known in the target domain.  The purpose is to improve the classification performance in the target domain by exploiting the source domain labels as well as some presumed relation between the two domains.

A variety of domain adaptation approaches have been proposed so far. Some methods are based on reweighing the samples for removing the sample selection bias \cite{HuangSGBS06, SunCPY11}. Another common line of solution is to align the source and the target domains in a joint feature space; via projections or transformations \cite{Fernando2013, GongSSG12, ZhangLO17}, kernel space representations \cite{PanTKY11, GhifaryBKZ17}, or deep networks \cite{LongC0J15, WangD18}. All such methods share the property that they strictly depend on feature space representations of data. The common effort in these works is to successfully align the source and the target distributions via transformations based on certain techniques and assumptions. Such transformations are relatively easy to compute when the deviation between the two distributions is small. On the other hand, when the source and the target distributions differ significantly, feature space methods often suffer from some performance loss as the intricacy of the actual transformation between the two domains is beyond the representation power of the transformation models they employ. Moreover, in certain machine learning problems involving classification or regression on platforms such as social networks, feature space representations may even not be available as data samples correspond to graph nodes (e.g. users), rather than vectors in an ambient space. 


\begin{figure}[t]
  \centering
  \centerline{\includegraphics[width=9.0cm]{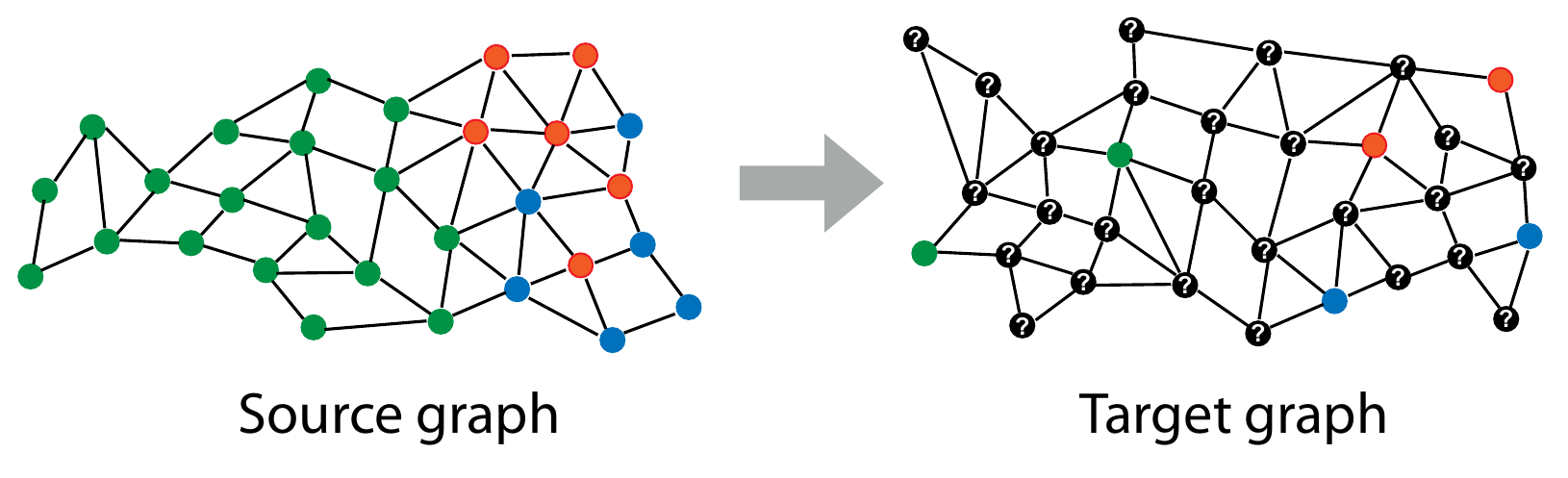}}
  \caption{Illustration of the domain adaptation problem studied in our work. A source graph with many labeled samples and a target graph with few labeled samples are illustrated. Using the source label information and the similarity between the local characteristics of the label functions on the two graphs, the unknown labels on the target graph are estimated.}\medskip
  \label{fig:illus_da_problem}
\end{figure}

In this work, we propose a domain adaptation method that aims to overcome such shortcomings. The proposed method is based purely on graph representations of the source and the target domains.  Graph methods provide flexibility in challenging setups where the two feature spaces are hard to align due to high dimensionality or the nonlinearity and irregularity of the warping between the two domains. In the proposed approach, the source and the target domains are represented respectively with a source graph and a target graph. We assume that few or no labels are available in the target domain. The computation of the class labels in the target domain is then cast as the estimation of an unknown label function on the target graph. Our assumption is that the variation of the class label function bears similar characteristics between the source and the target graphs: In a typical classification problem, different classes may have different characteristics regarding their localization, spread, and separability from each other. This is illustrated in Figure \ref{fig:illus_da_problem}, where the green class has wider spread than and better separation from the other two classes, while the red and the blue classes tend to be more localized and entangled with each other. Then, assuming that such class-specific characteristics are common between the source and the target domains, our purpose is to obtain an accurate estimate of the class labels on the target graph.

Our solution is based on the idea of representing the source and the target label functions in terms of a set of localized signals on the two graphs. The extension of classical signal processing techniques to graph domains has been an emerging research topic of the recent years, during which graph equivalents of concepts like harmonic analysis and filtering have been developed \cite{ShumanNFOV13, BronsteinBLSV17, DongTRF19}. In our work, we choose to represent class label functions in terms of graph wavelets \cite{HammondVG11}. The spectral graph wavelets proposed in \cite{HammondVG11} inherit their distinguishing characteristics such as adjustable scale and localization from the wavelets in classical signal processing theory. Graph wavelets provide convenient representations for our problem since we are interested in capturing class-specific variations of the label function in relation to, e.g., how localized each class is and how fast the label function tends to change in different regions of the graph.


\begin{figure}[t!]
  \centering
  \centerline{\includegraphics[width=8.5cm]{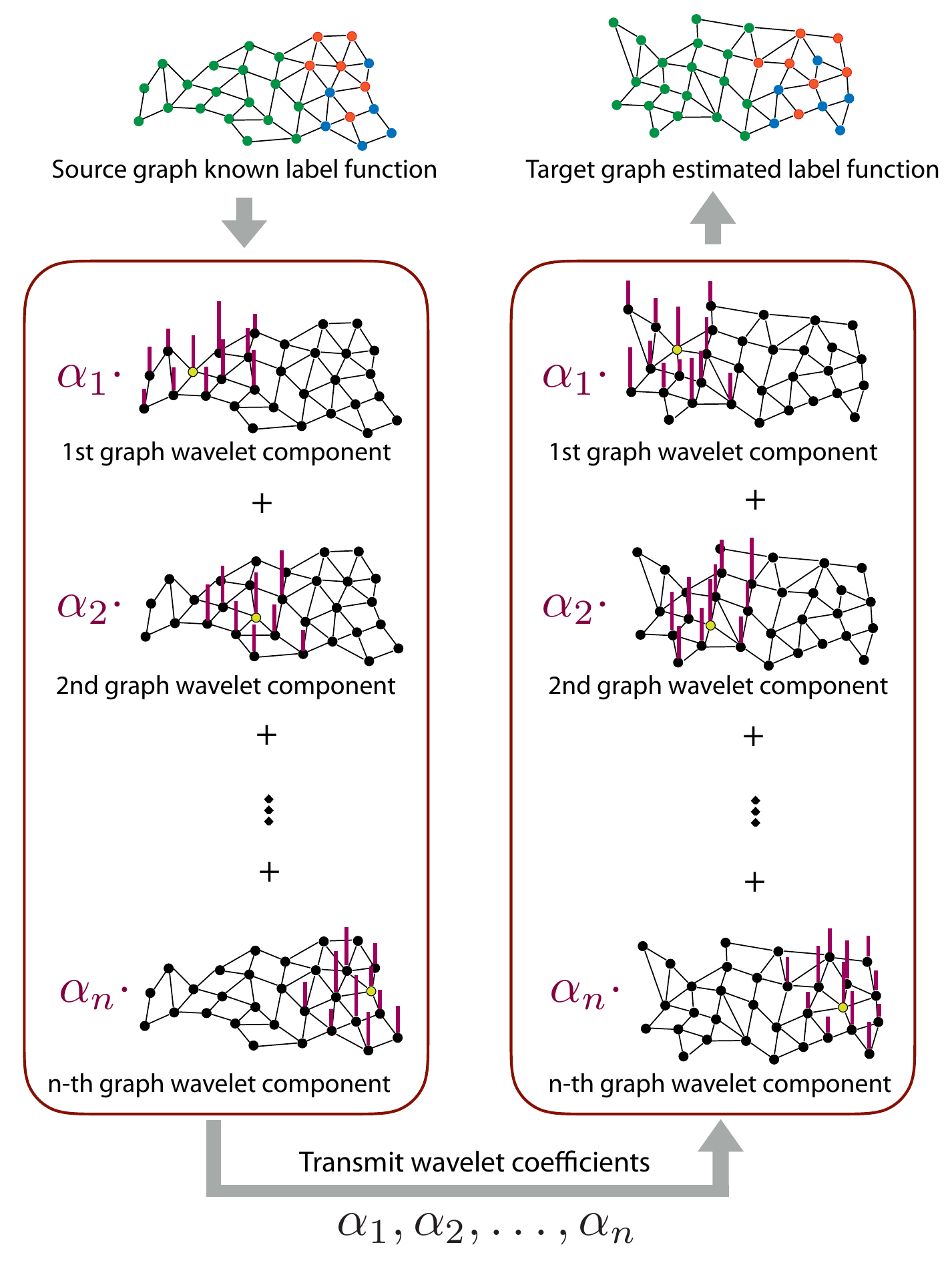}}
  \caption{Illustration of the proposed method. Graph wavelets are represented with bars, whose heights indicate the signal amplitude. Matched node pairs are shown with yellow color. The label function on the source graph is projected onto graph wavelets localized around matched nodes and the projection coefficients are transferred to the target graph. The label function on the target graph is estimated using the wavelet coefficients.}\medskip
  \label{fig:illus_da_alg}
\end{figure}

A mild assumption of our method is the availability of a small set of matches across the source and the target graphs, where a ``match'' refers to a source-target data sample pair belonging to the same class and related in a particular way depending on the problem, such as originating from related data resources. For instance, in a text classification problem, a pair of text feature vectors representing the same article written in a source language and translated into a target language may constitute a ``match'' between the two domains.  Similarly, in a web page classification problem, a ``match'' may consist of an image feature (source) and a text feature (target) that are extracted from the same web page. Although its reliance on the availability of a set of matches may seem to be a limitation of our framework, our algorithm often requires a small number of matches to attain satisfactory performance. Also,  in many practical settings, even if such matches are not available beforehand, it is often easy to form a small subset of such matches and inject them into the data set.

We formulate the connection between the source and the target domains via graph wavelets localized around the matched nodes across the two graphs, which serve as ``anchor'' points for characterizing and sharing the class-specific, local behaviors of the label functions. The information of the source label function is transmitted to the target graph through its projection onto the graph wavelets as depicted in Figure \ref{fig:illus_da_alg}. Our formal problem definition is then the following: Given a source graph and a target graph, we estimate the unknown labels such that the projections of the source and the target label functions onto the graph wavelets localized around matched graph nodes have similar coefficients. Employing also smoothness and label consistency priors on the label functions, the label estimation is formulated as a convex and quadratic optimization problem that can be solved analytically. In our method, the source and the target graphs are constructed independently of each other, in possibly different ambient spaces; therefore, the proposed method is conveniently applicable in heterogeneous domain adaptation settings. Also, as it does not strictly require the knowledge of labels in the target domain, it can be applied to unsupervised domain adaptation problems as well. Experimental results show that the proposed method performs successfully and achieves state-of-the-art classification accuracy on various types of data.

Our main contributions in this paper are the following.

\begin{itemize}
\item We formulate the domain adaptation problem in a pure graph setting, unlike mainstream methods that formulate the problem in a feature space.  
\item We propose the idea of representing label functions via spectral graph wavelets for the first time in the context of domain adaptation.
\item We present an algorithm that estimates the target label function by fitting its wavelet coefficients to those of the source label function.
\end{itemize}

The rest of the paper is organized as follows. In Section \ref{sec:rel_work}, we  overview the related literature. In Section \ref{sec:gda}, we present a brief introduction to basic concepts in spectral graph theory, overview spectral graph wavelets, and then describe the proposed domain adaptation algorithm. In Section \ref{sec:exp_res}, we evaluate the performance of the proposed method with comparative experiments. In Section \ref{sec:concl}, we conclude.

\section{Related Work}
\label{sec:rel_work}
	
Getting beyond the traditional machine learning paradigm that expects the training and test data to share similar characteristics, from late 90s on, researchers started studying whether one could efficiently transfer knowledge between two domains with differing distributions. Early works in this field explored relatively simple schemes such as sample reweighting, co-training, and basic feature augmentation solutions, while succeeding research efforts progressively shifted towards methods that actively seek to construct a mapping between the two domains \cite{PanY10}. Currently, a great body of works focuses on learning elaborate and rich models from large data sets with sophisticated techniques \cite{WangD18}.

The most basic form of domain adaptation is the covariate shift or sample selection bias problem, where the conditional distribution of the labels is assumed to be the same between the source and the target domains. Sample reweighing approaches can be successfully employed in this setting \cite{HuangSGBS06, SunCPY11}. In the more typical case where the conditional distributions are different, a relatively simple solution consists of mapping the source and the target data to a common high dimensional space via feature augmentation \cite{DaumeKS10, DuanXT12}. The semi-supervised EA++ method proposed in \cite{Daume2010} augments the source and the target features into a higher dimensional space where the source and target hypotheses are made to agree on unlabeled target data. Some works have extended the domain adaptation problem to settings with more than one source domain, where a target classifier is computed from the classifiers in the source domains \cite{CrammerKW08, WuWZTXYH17}. 

	
A quite prevalent approach in the literature is to align the two domains using a transformation or a projection \cite{Fernando2013, GongSSG12, ZhangLO17, PereiraT18, LiangRZT19}. The SA algorithm proposed in \cite{Fernando2013} is an unsupervised method that aligns the source and target PCA bases by learning a linear transformation between the two domains. The unsupervised GFK algorithm in \cite{GongSSG12} is based on a similar subspace alignment idea; however, the source and target bases are modeled as two distinct points on a Grassmann manifold and the optimal transformation between the two bases is found by computing the geodesic curve joining them on the manifold. The JGSA algorithm in \cite{ZhangLO17} computes projections of the source and target data into a common domain by maximizing the target scatter and the between-class source scatter, minimizing the within-class source scatter, and also minimizing the divergence between the two distributions at the same time. The SCA algorithm in \cite{GhifaryBKZ17} similarly computes projections by optimizing the between-class and within-class scatters in the two domains along with the domain scatters and the total scatter; however, formulates the problem in a Reproducing Kernel Hilbert Space.

Some domain adaptation methods aim to directly match the densities or the covariances of the two distributions in order to align the two domains  \cite{LopezPazHS12, SunFS16}, while others propose a solution based on learning a metric  \cite{XuPXWLMS17, HerathHP17} or sparse representations \cite{TaoHW14, YangMY18}. In some works, the learning of a classifier is incorporated into the problem of learning a mapping \cite{YaoPNLM15}. The recent LDADA method proposed in \cite{LuSC0H18} learns a classifier in the original data space via self-training, where the scatter matrices computed from class means are used for identifying optimal projection directions. The extraction of domain-invariant features via deep networks has also been an active research topic of the last few years  \cite{LongC0J15, WangD18}. A common strategy is to reduce the deviation between the source and the target features via adversarial learning, where the end classifiers are also often jointly optimized with the feature extraction layers \cite{LongZ0J17, TzengHSD17}.


While many approaches focus on feature-space representations of data as discussed above, there are also methods that incorporate a graph model in the learning. Several algorithms use smoothness priors of the label function on the data graph \cite{ChengP14, XiaoG15} . However, there are much fewer examples of methods explicitly making use of  graph bases or dictionaries for representing graph signals in classification problems as in our work, even outside the context of domain adaptation. The studies in  \cite{EynardKBGB15, RodolaCBTC17} employ graph Fourier bases in solving multiview 3D shape analysis or clustering problems. The methods in \cite{ThanouSF14, ThanouF18}  propose to use sparse signal representations on graphs via localized graph dictionaries; however, in an unsupervised setting where the purpose is to reconstruct and approximate graph signals. Finally, the recent study \cite{PilanciV20} proposes to represent label functions over graph Fourier bases, which is similar to our work in the sense that it treats the domain adaptation problem in a pure graph environment. The label function on the target graph is estimated based on the similarity of its Fourier coefficients to those on the source graph, while jointly learning a transformation that aligns the two graph Fourier bases.

\section{Graph Domain Adaptation with Localized Signal Representations}	
\label{sec:gda}
	
In this section, we present our graph domain adaptation method. We first overview spectral graph theory  and spectral graph wavelets in Sections \ref{ssec:sgt} and \ref{ssec:sgw}. We then describe our domain adaptation algorithm based on transferring wavelet coefficients between the source and the target graphs in Section \ref{ssec:prop_method}.

\subsection{Overview of Spectral Graph Theory}
\label{ssec:sgt}
	
Here, we briefly overview basic concepts from spectral graph theory \cite{Chung97} and graph signal processing \cite{ShumanNFOV13}. Throughout the paper, matrices and vectors are represented with uppercase and lowercase letters.  Let $\G=(\V, \E, W)$ be a graph consisting of $\N$ vertices (nodes) in the vertex set $\V=\{ x_i \}_{i=1}^\N$ and edges $\E$, where the matrix $W$ stores the edge weights. A graph signal is a function $f: \V \rightarrow \R$ taking a real value on each graph node $x_i$. A graph signal can equivalently be regarded as a vector $f \in \R^{\N}$ in the $\N$-dimensional space, which we adopt in our notation.

The weight matrix $W\in \R^{\N \times \N}$ is a symmetric matrix consisting of nonnegative edge weights, such that $W_{ij} $ is the weight of the edge between the nodes $x_i$ and $x_j$. If the nodes $x_i$ and $x_j$ are not connected with an edge, then $W_{ij}=0$.   The weight matrix $W$ defines a diagonal degree matrix $D \in \R^{\N \times \N}$ given by $D_{ii} = \sum_j W_{ij}$. The graph Laplacian $L \in \R^{\N \times \N}$ is then defined as\footnote{Note that the normalized version $L=D^{-1/2}(D-W)D^{-1/2}$ of the graph Laplacian may also be preferred in some settings.}
	\begin{equation}
	\label{eq:defn_graph_Lap}
	L=D-W,
	\end{equation}
which can be seen as an operator acting on a function $f$ via the matrix multiplication $Lf$. The graph Laplacian $L$ is of crucial importance in graph signal processing since it allows the extension of concepts such as Fourier transform and filtering to graph domains.  Several previous studies have shown that the graph Laplacian $L$ can be regarded as the graph equivalent of the well-known Laplace operator in the Euclidean domain, or the Laplace-Beltrami operator on manifold domains \cite{ShumanNFOV13}, \cite{HeinAv05}.


The Laplace operator $\Delta$ in classical signal processing has the important property that complex exponentials $e^{j\Omega t}$ used in the definition of the Fourier transform are given by its eigenfunctions
$
- \Delta(e^{j\Omega t}) = \Omega^2 e^{j\Omega t}
$.
In analogy with this property in classical signal processing, the eigenvectors $u_1, \dots, u_\N \in \R^{\N}$ of the graph Laplacian satisfying
$
	L u_k = \lambda_k u_k
$
	for $k=1, \dots, \N$ are of special interest since they define a Fourier basis over graph domains \cite{ShumanNFOV13}. Indeed, for any graph, the eigenvector $u_1$ corresponding to the smallest eigenvalue $\lambda_1=0$ is always a constant function on the graph, while the speed of variation of $u_k$ on the graph  increases  for increasing $k$. Hence, the eigenvalues $\lambda_1, \dots, \lambda_\N$ of the graph Laplacian correspond to frequencies such that $\lambda_k$ gives a measure of the speed of variation of the eigenvector $u_k$ regarded as a Fourier basis vector. 
		
	The definition of the graph Fourier basis allows the extension of the Fourier transform to graph domains as follows. Given a graph signal $f \in \R^\N$, its Fourier transform $\hat f$ is defined as
	\begin{equation}
	\label{eq:defn_Fourier_tran}
	\hat f (\lambda_k) = \langle  f , u_k \rangle,
	\end{equation}
	such that the $k$-th Fourier coefficient $\hat f (\lambda_k)$ is given by the inner product of $f$ and the Fourier basis vector $u_k$. Throughout the paper, $\lambda$ stands for a frequency variable, and the Fourier transform $\hat f$ is defined on the frequencies $\lambda_1, \dots, \lambda_\N$ determined by the graph topology, as common convention in graph signal processing \cite{ShumanNFOV13}. The inverse Fourier transform similarly corresponds to the reconstruction of the graph signal $f$ as the sum of the Fourier basis vectors weighted by the Fourier coefficients 
$
	f =  \sum_{k=1}^\N \hat f (\lambda_k) u_k 
$.

	Based on these definitions, one can generalize the filtering operation to graph domains as well. Given a filter kernel $g(\lambda)$ specified in the spectral domain as a function of the frequency variable $\lambda$, an ``input'' graph signal $f$ can be filtered by multiplying its Fourier transform with the filter kernel as
	$
	\hat h (\lambda_k) = \hat f(\lambda_k) g(\lambda_k)
$,
where $\hat h (\lambda_k)$ is the Fourier transform of the ``output'' signal after filtering. This spectral representation can then be transformed back to the vertex domain via the inverse Fourier transform in order to obtain the output signal $h$ as
	\[
	h = \sum_{k=1}^N \hat f(\lambda_k) g(\lambda_k) u_k .
	\]

	\subsection{Spectral Graph Wavelets}
	\label{ssec:sgw}

	The wavelet transform is a widely used transform in many signal processing applications such as compression and reconstruction \cite{VetterliK95}. In the recent years, several extensions of the wavelet transform have been proposed for graph domains \cite{HammondVG11, CoifmanM06, NarangO12}. In our graph domain adaptation problem, we would like to efficiently capture and transfer the local characteristics of label functions on graphs such as their vertex spread and speed of change. For this reason, in our work we prefer to use spectral graph wavelets \cite{HammondVG11} for representing label functions, which are theoretically shown to enjoy desirable properties such as good localization in the vertex domain and the spectral domain.
	
	Spectral graph wavelets \cite{HammondVG11} are defined and characterized essentially in the spectral domain, based on the idea of extending the traditional wavelet transform to graphs. Considering a graph $\G$ with $\N$ nodes, the spectral graph wavelet transform is specified by a kernel $g(\lambda)$ in the spectral domain, which is a function of the frequency variable $\lambda$. The wavelet kernel $g(\lambda)$ represents a band-pass filter and acts on a graph signal $f$ in the frequency domain via its Fourier transform $\hat f$ as
$
	\hat T_g f (\lambda_k) = g(\lambda_k ) \hat f (\lambda_k)
$.
	This operation corresponds to band-pass filtering the signal $f$ to obtain a new signal $ T_g f $. The graph signal $T_g f \in \R^\N$ can be reconstructed in the vertex domain by taking the inverse Fourier transform of $\hat T_g f $ as
	\[
	T_g f  = \sum_{k=1}^N   \hat T_g f (\lambda_k) \, u_k =  \sum_{k=1}^N   g(\lambda_ k) \hat f (\lambda_k)  u_k,
	\]
	where $u_1, \dots, u_N$ are the Fourier basis vectors. 
	
\begin{figure}[t]
		\centering
		\begin{subfigure}[t]{0.23\textwidth}
			\centering
			\includegraphics[height=2.3cm]{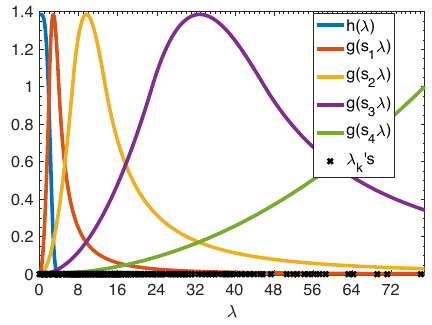}
			\caption{}
		\label{fig_wavelet_visual_source_a}
		\end{subfigure}
		~
		\centering
		\begin{subfigure}[t]{0.23\textwidth}
			\centering
			\includegraphics[height=2.3cm]{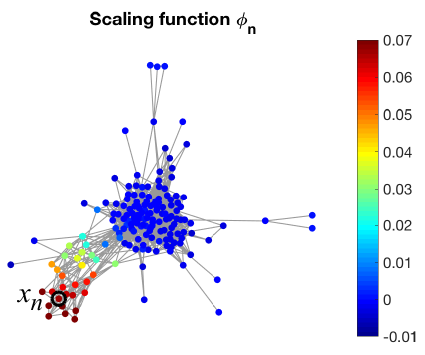}
			\caption{}
		\label{fig_wavelet_visual_source_b}
		\end{subfigure}	
		~
		\centering
		\begin{subfigure}[t]{0.23\textwidth}
			\centering
			\includegraphics[height=2.3cm]{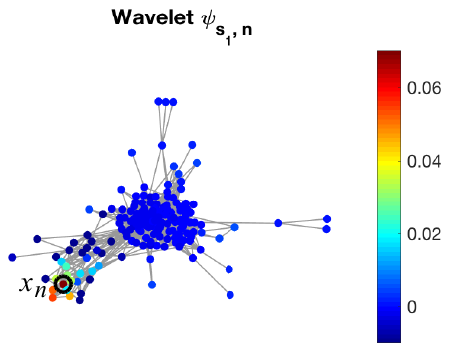}
			\caption{}
		\label{fig_wavelet_visual_source_c}
		\end{subfigure}
		~
		\centering
		\begin{subfigure}[t]{0.23\textwidth}
			\centering
			\includegraphics[height=2.3cm]{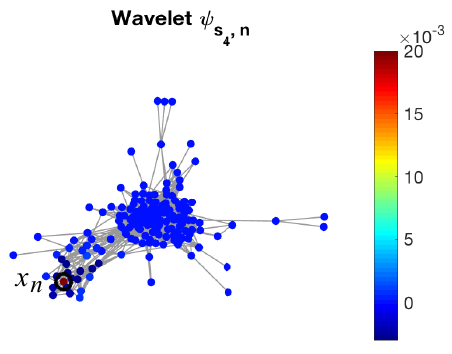}
			\caption{}
		\label{fig_wavelet_visual_source_d}
		\end{subfigure}					
		\caption{Spectral graph wavelets formed on a graph extracted from the Facebook network \cite{leskovec2012learning}. Panel (a) shows the wavelet kernel $g(s\lambda)$ scaled at four different scale values $s_1$, $s_2$, $s_3$, and $s_4$; the scaling function kernel $h(\lambda)$; and the eigenvalues $\lambda_1, \dots, \lambda_N$ of the graph. Panels (b)-(d) show the scaling function and the wavelets at scale values $s_1$ and $s_4$, localized at the node $x_n$ marked with the black circle. Signal amplitudes are indicated with color codes such that the amplitude increases as the color shifts from blue to red.}
       \label{fig:wavelet_visual}
	\end{figure}

	This filtering operation is used for defining the spectral graph wavelets as follows. First, in order to define the graph wavelet transform at an arbitrary scale $s$,  the wavelet kernel is simply scaled as $g(s \lambda)$ as illustrated in Figure \ref{fig_wavelet_visual_source_a}. Then, the spectral graph wavelet vector $\psi_{s,n} \in \R^\N  $ localized at node $x_n$ and having scale $s$ is obtained by band-pass filtering the Dirac delta function $\delta_n \in \R^\N$ localized at node $x_n$ (which is the graph signal taking the value 1 at node $x_n$ and 0 elsewhere) using the filter kernel $g(s \lambda)$
	\begin{equation}
	\label{eq:defn_wavelet}
	\psi_{s,n} = T_g^s \delta_n =  \sum_{k=1}^N   g(s \lambda_ k ) \hat \delta_n (\lambda_k)  u_k,
	\end{equation}
	which is illustrated in Figures \ref{fig_wavelet_visual_source_c}-\ref{fig_wavelet_visual_source_d}. Here $T_g^s$ denotes the filtering operation with kernel $g$ at scale $s$; and $ \hat \delta_n $ stands for the graph Fourier transform of the Dirac delta function $\delta_n$ as given by \eqref{eq:defn_Fourier_tran}. Hence, the definition in \eqref{eq:defn_wavelet} indicates how one can generate a collection of graph wavelet functions $\psi_{s,n} $ at different scales $s$ and localized at different graph nodes $x_n$, where the exact structure of the wavelets are specified by the filter kernel $g(\lambda)$, similarly to the way that wavelets are obtained from mother wavelet kernels in traditional signal processing. Having defined the band-pass wavelet functions $\psi_{s,n}$ based on the band-pass kernel $g(\lambda)$, one can similarly start with a low-pass kernel $h(\lambda)$ and use it to define the scaling function $\phi_{n} \in \R^\N $ localized at node $x_n$, which is a low-pass graph signal just as in classical wavelet analysis  \cite{HammondVG11}. The low-pass kernel $h(\lambda)$ and the scaling function $\phi_{n}$  are demonstrated in Figures \ref{fig_wavelet_visual_source_a} and \ref{fig_wavelet_visual_source_b}. It can be observed that the scaling function $\phi_{n}$ is a smooth and slowly varying function on the graph, in line with the low-pass structure of the kernel $h(\lambda)$. On the other hand, while both wavelets in Figures \ref{fig_wavelet_visual_source_c} and \ref{fig_wavelet_visual_source_d} exhibit band-pass characteristics, $\psi_{s_4,n}$ has a faster variation and changes more abruptly than $\psi_{s_1,n}$, as its spectrum is shifted much more towards higher frequencies.

	Finally, the wavelet coefficient of a signal $f$ at scale $s$ and localized at node $x_n$ is found by taking the inner product $\langle   \psi_{s,n} , f   \rangle$  between $f$ and the wavelet $\psi_{s,n}$. 
	%
%
We refer the reader to \cite{HammondVG11} for a detailed analysis of the localization properties of graph wavelets and scaling functions and also on what constitutes a proper choice of the kernels $g(\lambda)$ and $h(\lambda)$ in order to ensure such properties. It is further discussed in  \cite{HammondVG11} that, under an appropriate sampling $s_1, \dots, s_J$ of the scale parameter $s$, the scaling function and the wavelets $\{ \phi_n \} \cup \{ \psi_{s_j, n} \}$, $n=1, \dots, N, \ j= 1, \dots, J$ form a frame. In our work, we use such a collection of graph signals, which consists of a low-pass scaling function and a couple of band-pass wavelets at different nodes. In the next section, we discuss how one can exploit these localized signal representations in order to transfer knowledge from one graph to another in domain adaptation.

	\subsection{Domain Adaptation with Spectral Graph Wavelets}
	\label{ssec:prop_method}
	
	We now focus on the problem of domain adaptation in a setting with a source graph $\G^s=(\V^s, \E^s, W^s)$ and a target graph $\G^t=(\V^t, \E^t, W^t)$ whose nodes represent the source data $\{ x^s_i \}_{i=1}^{\Ns}$ and the target data $\{ x^t_i \}_{i=1}^{\Nt}$. In some settings, graphs are known and available beforehand, e.g. as in problems related to social networks, whereas in some other applications, the graphs are to be constructed from the data set. While the construction of the graphs is out of the scope of our work, it is often suitable to adopt common strategies such as choosing the neighbors with respect to the K-NN rule and setting the edge weights with a Gaussian kernel. Let $L^s \in \R^{\Ns \times \Ns}$ and $L^t \in \R^{\Nt \times \Nt}$ denote the source and the target graph Laplacian matrices obtained from the weight matrices $W^s \in \R^{\Ns \times \Ns}$ and $W^t \in \R^{\Nt \times \Nt}$ as in \eqref{eq:defn_graph_Lap}.    
	
	We consider a  pair of class label functions $f^s \in \R^{\Ns}$ and $f^t \in \R^{\Nt}$, respectively on the source and the target graphs. We address a setting where the class labels are largely  known on the source graph, whereas few labels are known on the target graph. Let $y^s_i$ denote the label of the source node $x^s_i$ whenever $x^s_i$ is labeled, and let $y^t_i$ be defined similarly in the target domain. We assume that a small set of matched nodes $\{ (x^s_{m_i}, x^t_{n_i}) \}_{i=1}^\K$ between the source and the target graphs is available\footnote{An arbitrary indexing with the indices $m_i$ and $n_i$ is preferred in our notation since no relation is assumed to be known between the source and the target graphs apart from the matched node pairs. The node enumerations are assumed to be arbitrary.}.  These matched node pairs serve as anchor points in transferring the knowledge of the local behavior of the label functions $f^s$ and $f^t$ between the two graphs. In practice, these node pairs may consist of data samples related in a certain way, e.g., extracted from the same resource.
	
	Our method is based on the following idea: In a setting where the label functions $f^s$ and $f^t$ share similar local characteristics on the source and the target graphs, the wavelet coefficients of the label functions must also be similar. For a pair of matched nodes $ (x^s_{m_i}, x^t_{n_i})$ known to be related to each other, this assumption can be formulated as  
	\begin{equation*}
	\langle   \psi^s_{s, m_i} , f^s   \rangle \approx \langle   \psi^t_{s,n_i} , f^t   \rangle 	
	\quad \text{ and } \quad
	\langle   \phi^s_{m_i} , f^s   \rangle \approx \langle   \phi^t_{n_i} , f^t   \rangle	
	\end{equation*}
	at all scales $s$. Here $\psi^s_{s,m_i}$ and $\phi^s_{m_i}$ respectively denote the wavelet at scale $s$ and the scaling function on the source graph localized at node $x^s_{m_i}$, and  similarly $\psi^t_{s,n_i}$ and $\phi^t_{n_i}$ denote those on the target graph localized at node  $x^t_{n_i}$. One may wonder about the validity of this assumption; i.e., whether the wavelet coefficients may indeed be expected to match in this way in practice, especially considering that the source and the target graphs are constructed independently. While there already exist results showing that wavelet functions constructed on topologically similar graphs are close to each other \cite{DonnatZHL18}, we further study our assumption in Section \ref{ssec:comp_wavelet_coef} and experimentally confirm that the projections of the label functions onto the wavelets on the source and the target graphs yield quite similar coefficients in typical domain adaptation problems.

	Let $\Psi^s_{m_i} \in \R^{\Ns \times (J+1)}$ and $\Psi^t_{n_i} \in \R^{\Nt \times (J+1)}$ denote respectively the matrices whose columns consist of the source and the target scaling functions and wavelets for a sampling $\{s_1, \dots s_J\}$ of the scale parameter at a matched pair $(x^s_{m_i}, x^t_{n_i}) $
	\begin{equation}
	\begin{split}
	\Psi^s_{m_i}=[ \phi^s_{m_i} \ \psi^s_{s_1, m_i} \ \dots \ \psi^s_{s_J, m_i}  ]  ,
	\quad \quad
	\Psi^t_{n_i}=[ \phi^t_{n_i} \ \psi^t_{s_1, n_i} \ \dots \ \psi^t_{s_J, n_i}  ]  .
	\end{split}
	\end{equation}
	In the determination of the wavelet functions, we adopt a logarithmic sampling of the scale parameter $s$ as proposed in \cite{HammondVG11}. Let us also define the matrices $\Psi^s \in  \R^{\Ns \times \K(J+1)}$ and $\Psi^t \in \R^{\Nt \times \K(J+1)}$ containing the scaling functions and wavelets at all $\K$ matched node pairs as
	\begin{equation}
	\label{eq:form_wavelet_dict}
	\begin{split}
	\Psi^s=[ \Psi^s_{m_1} \  \Psi^s_{m_2}  \  \dots  \Psi^s_{m_\K}]  , 
	\quad \quad
	\Psi^t=[ \Psi^t_{n_1} \  \Psi^t_{n_2}  \  \dots  \Psi^t_{n_\K}]  .
	\end{split}
	\end{equation}
	%


	In a setting where labels are completely available on the source graph, evaluating the graph wavelet coefficients of the source label function $f^s$ at the source matched nodes $\{ x^s_{m_i} \}$ and transferring them to the target graph along their correspondences $\{ x^t_{n_i} \}$, one can reconstruct the target label function based on the transferred wavelet coefficients. If a reliable set of sufficiently many matches are available, one may expect to have a good reconstruction of $f^t$ on the target graph by solving the inverse problem
	$
	(\Psi^t)^T f^t  = (\Psi^s)^T f^s 
	$.
	However, some possible challenges that might be encountered are the following. First, the number of matches might be limited in practice.  Moreover, one may not always have a complete set of labels in the source domain, in which case the source wavelet coefficients $ (\Psi^s)^T f^s  $ cannot be computed. Due to these difficulties, we propose to estimate the label functions $f^s$ and $f^t$ by solving the following optimization problem
	\begin{equation}
	\label{eq:wda_opt_prob}
	\begin{split}
	\min_{f^s, f^t} F(f^s, f^t),   
	\end{split}
	\end{equation}
where
	\begin{equation}
	\label{eq:defn_obj_func}
	\begin{split}
	F(f^s, f^t) &=  \| S^s f^s - y^s \|^2 +  \| S^t f^t - y^t \|^2 + \mu \| (\Psi ^s)^T f^s -  (\Psi ^t)^T f^t  \|^2  \\
	+ & \gamma_s ( (f^s)^T L^s f^s )  +  \gamma_t ( (f^t)^T L^t f^t ).  
	\end{split}
	\end{equation}
Here, $y^s \in R^{\Ms}$ and  $y^t \in R^{\Mt}$ are label vectors consisting of the known labels $\{ y^s_i \}$ and $\{ y^t_i \}$ on the source and the target graphs, where $\Ms$ and $\Mt$ are the number of known labels on each graph. The matrices $S^s  \in \R^{\Ms \times \Ns}$ and $S^t \in \R^{\Mt \times \Nt} $ are binary selection matrices consisting of $0$'s and $1$'s, which extract the labeled entries of $f^s$ and $f^t$ and enforce them to be in agreement with the known labels in the vectors $y^s$ and $y^t$. The third term in \eqref{eq:defn_obj_func} imposes the source wavelet coefficients  $(\Psi^s)^T f^s$ to be similar to the target wavelet coefficients  $(\Psi^t)^T f^t$ at the set of matched nodes. Finally, the last two terms serve as regularization terms that prevent the label function estimates  $f^s$ and $f^t$ from varying too fast on the graphs, via the source and the target graph Laplacians $L^s$ and $L^t$. The parameters $\mu$, $\gamma_s$, and $\gamma_t$ are nonnegative scalars weighting the contribution of each term. 
	
	In order to solve the optimization problem in \eqref{eq:wda_opt_prob}, we first notice that the objective function $F(f^s, f^t)$ is convex with respect to the optimization variables $f^s$ and $f^t$, since the graph Laplacian matrices $L^s$ and $L^t$ are always positive semi-definite. 
	We can then simply solve the problem by analytically setting the derivatives of $F(f^s, f^t)$ with respect to $f^s$ and $f^t$ to 0 as
	\begin{equation}
	\begin{split}
	\frac{\partial F(f^s, f^t)}{\partial f^s} =& 2 (S^s)^T S^s f^s - 2 (S^s)^T y^s 
	+ 2 \mu \Psi^s (\Psi^s)^T f^s - 2 \mu \Psi^s (\Psi^t)^T f^t  \\
	&+ 2 \gamma_s L^s f^s =0 \\
	\frac{\partial F(f^s, f^t)}{\partial f^t} =& 2 (S^t)^T S^t f^t - 2 (S^t)^T y^t
	+ 2 \mu \Psi^t (\Psi^t)^T f^t - 2 \mu \Psi^t (\Psi^s)^T f^s  \\
	&+ 2 \gamma_t L^t f^t =0. \\
	\end{split}
	\end{equation}
	Solving these two equations together, we get 
	\begin{equation}
	\label{eq:comp_ft_fs}
	\begin{split}
	f^t = &((S^t)^T S^t  + \mu  \Psi^t (\Psi^t)^T   - \mu^2 \Psi^t (\Psi^s)^T A \Psi^s (\Psi^t)^T  + \gamma_t L^t  )^{-1}  \\
	&((S^t)^T y^t  + \mu  \Psi^t (\Psi^s)^T A  (S^s)^T y^s ) \\
	f^s = &A  ( (S^s)^T y^s + \mu \Psi^s (\Psi^t)^T  f^t ),
	\end{split}
	\end{equation}
	where 
	\begin{equation}
	\label{eq:defn_A}
	A= ((S^s)^T S^s   + \mu \Psi^s (\Psi^s)^T  + \gamma_s L^s)^{-1}.
	\end{equation}
	This gives the estimates of the source and the target label functions $f^s$ and $f^t$. We call the proposed algorithm Graph Domain Adaptation with Matched Local Projections (GrALP) and summarize it in Algorithm \ref{alg:GrALP_Algorithm}.

\begin{algorithm}[t]
\caption{Graph Domain Adaptation with Matched Local Projections (GrALP)}

\begin{algorithmic}[1]

\STATE
\textbf{Input:} \\
$L^s$, $L^t$: Source and target graph Laplacian matrices\\
$y^s$, $y^t$: Available source and target labels\\

\textbf{Algorithm:} \\

\STATE 
Compute the wavelet matrices $\Psi^s$ and $\Psi^t$  as in \eqref{eq:form_wavelet_dict}.

\STATE
Compute the matrix $A$ as in \eqref{eq:defn_A}.

\STATE
Compute $f^s$ and $f^t$ as in \eqref{eq:comp_ft_fs}.

\STATE
\textbf{Output}:\\

$f^t $: Estimated target label function\\
$f^s $: Estimated source label function

\end{algorithmic}
\label{alg:GrALP_Algorithm}
\end{algorithm}

	\subsection{Complexity Analysis of the Algorithm}

 The complexity of the proposed method is analyzed in this section. 
First, assuming that the graph weight matrices $W^s$ and $W^t$ are known, the time complexities of computing the source and the target graph Laplacians $L^s$ and $L^t$ are of $O(\Ns^2)$ and $O(\Nt^2)$. Noting that the complexities of calculating the eigenvalue decompositions of $L^s$ and $L^t$ are respectively of $O(\Ns^3)$ and  $O(\Nt^3)$, the computations of the source and the target wavelet matrices $\Psi^s$ and $\Psi^t$ have complexities of $O(\Ns^2 J \K+\Ns^3)$ and $O(\Nt^2  J \K+\Nt^3)$. 

From \eqref{eq:defn_A} we observe that the $A$ matrix  can be computed with $O(\Ns^2 \Ms+\Ns^2J \K+\Ns^3)$ operations. Since the number of labeled samples $\Ms$ is always smaller than or equal to $\Ns$, this complexity reduces to $O(\Ns^2J \K+\Ns^3)$.  The complexity of computing the matrix inverse
%
%
in \eqref{eq:comp_ft_fs} can be found as $ O(\Nt^2 \Mt +\Nt^2 J \K+\Ns \Nt J \K+\Ns^2 \Nt+ \Nt^2 \Ns +\Nt^3)$. Since $\Mt \leq \Nt$, this complexity reduces to $O(\Nt^2 J \K+\Ns \Nt J \K+\Ns^2 \Nt+ \Nt^2 \Ns+ \Nt^3)$.

Next, the matrix 
$ ((S^t)^T y^t  + \mu  \Psi^t (\Psi^s)^T A  (S^s)^T y^s )
$ 
in \eqref{eq:comp_ft_fs} can be calculated with additional $O(\Nt \Mt + \Nt \Ns + \Ns \Ms)$ operations, after which $f^t$ can be found with $O(\Nt^2)$ operations.

The overall complexity of finding the label functions $f^s$ and $f^t$ is then of $O(\Ns^3+\Nt^3+ \Ns^2 J \K+ \Nt^2 J \K+\Ns \Nt J \K)$. Since the number of matches $\K$ is typically a small number,  we  may assume that $J \K$ is less than $\Ns$ or $\Nt$. The overall complexity can then be simplified as $O(\Ns^3+\Nt^3)$.


\section{Experimental Results}
\label{sec:exp_res}
	
	In this section, we first present the data sets used in the experiments, then evaluate the performance of the proposed GrALP method with comparative experiments, and finally analyze the sensitivity of the method to the selection of algorithm hyperparameters and wavelet kernels.
	
\subsection{Data sets and setting}	

The experiments are conducted on four real data sets. In data sets where the graphs are not readily available, a $K$-NN graph is constructed in each domain by connecting samples to their $K$ nearest neighbors with respect to the Euclidean distance. The edge weights are set with a Gaussian kernel as $W_{ij}=e^{- \| x_i - x_j \|^2 / (2\sigma^2)}$. The graph wavelets are constructed independently on the source and the target graphs. The wavelet kernel $g(\lambda)$ and the scaling function kernel $h(\lambda)$ in Section \ref{ssec:sgw} are respectively chosen as the AB-spline function and the low-pass kernel of the form $e^{-O(\lambda^4)}$ proposed in \cite{HammondVG11}. The number of wavelets is selected as $J=4$, and the wavelet scales $\{s_1, \dots, s_4\}$ are obtained with the logarithmic sampling strategy in \cite{HammondVG11}. The following data sets are used.

\begin{figure}[t]
		\centering
		\begin{subfigure}[t]{0.30\textwidth}
			\centering
			\includegraphics[width=3.5cm]{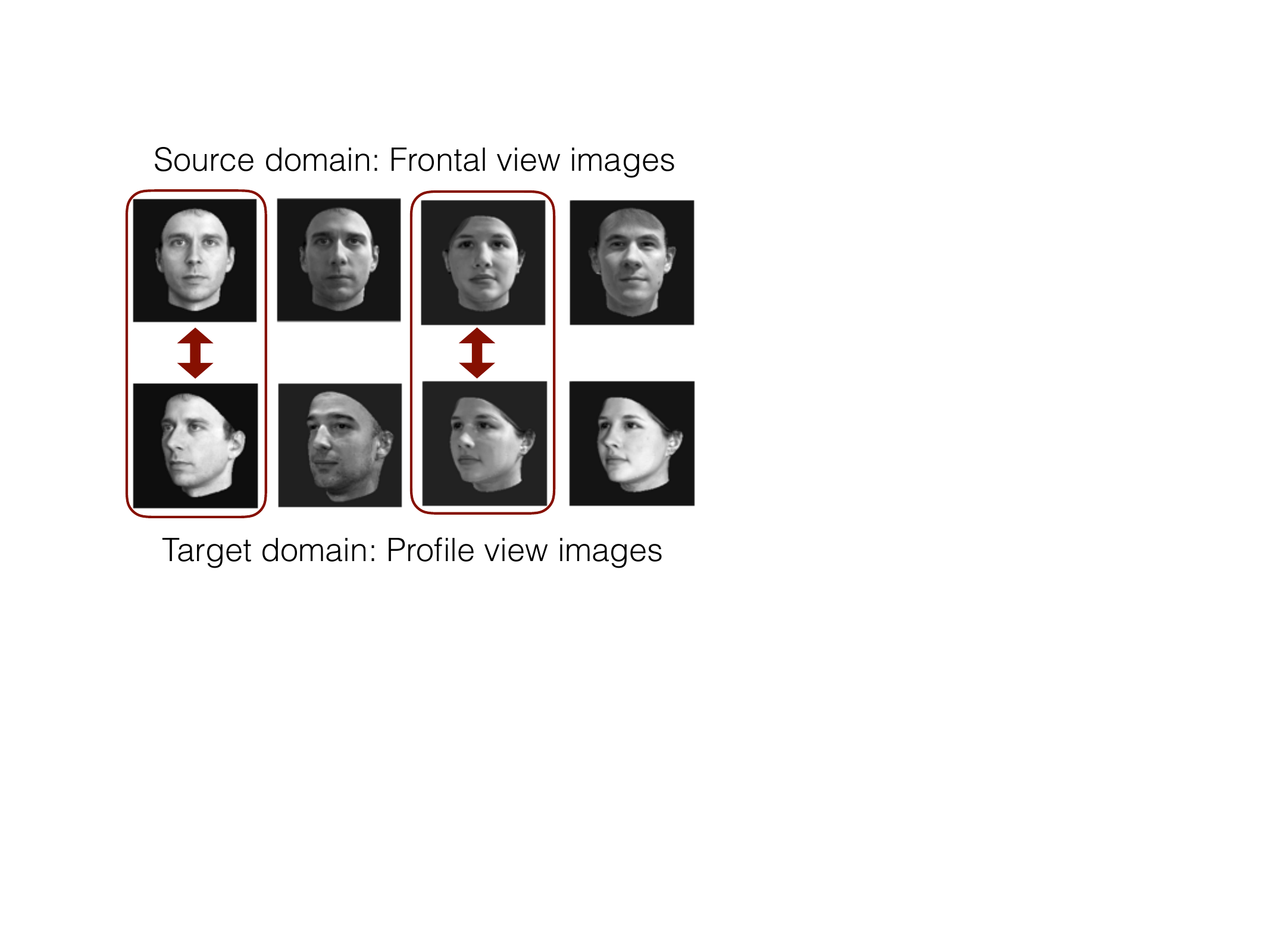}
			\caption{MIT-CBCL}
		\label{fig:sample_images_mit}
		\end{subfigure}
		~
		\centering
		\begin{subfigure}[t]{0.33\textwidth}
			\centering
			\includegraphics[width=4.4cm]{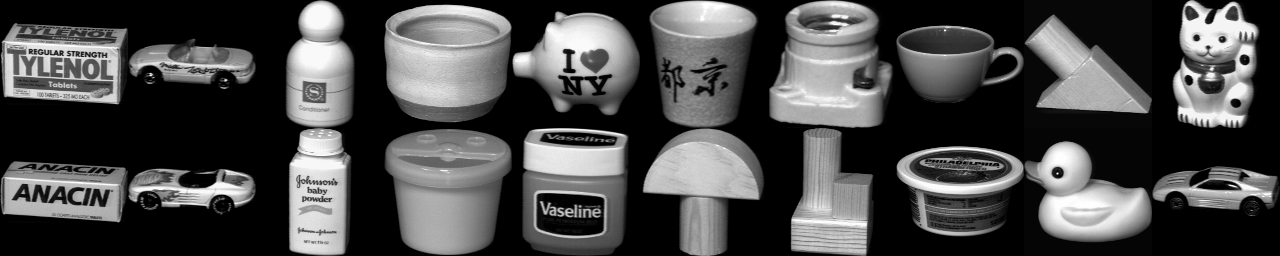}
			\caption{COIL-20}
		\label{fig:COIL20_classes}
		\end{subfigure}	
		~
		\centering
		\begin{subfigure}[t]{0.30\textwidth}
			\centering
			\includegraphics[width=4cm]{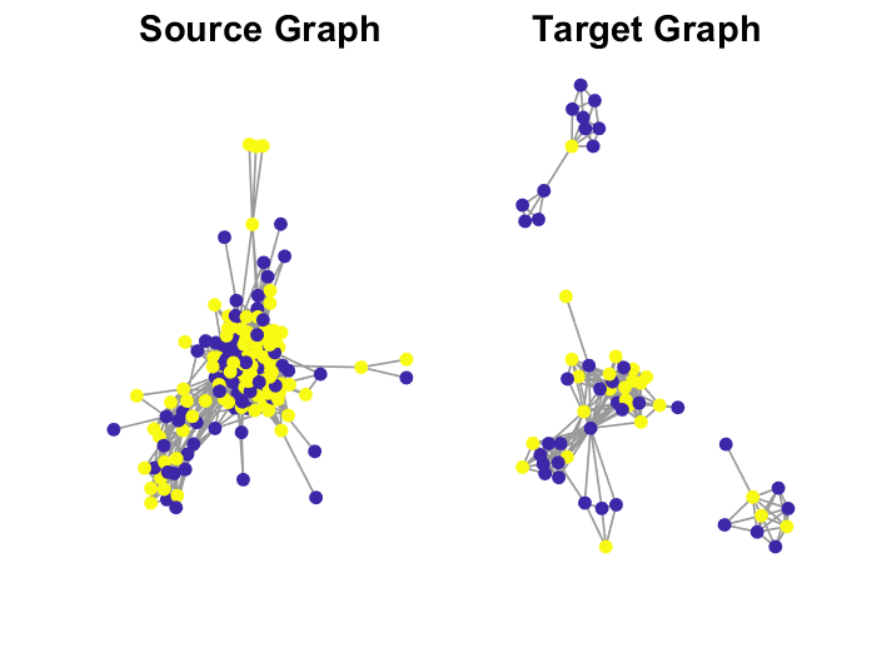}
			\caption{Facebook}
		\label{fig:Facebook_dataset}
		\end{subfigure}					
		\caption{Data sets used in the experiments. (a) Sample images from the MIT-CBCL data set. Arrows indicate  matched node pairs. (b) Sample images from the COIL-20 object data set. The top and the bottom rows represent the source and the target domains, respectively. Each column is an object pair assigned the same class label in the experiments. (c) The source and the target graph communities in the Facebook data set. The binary label function is illustrated with two different colors. }
	\end{figure}

\textbf{MIT-CBCL face image data set.}	 The MIT-CBCL data set \cite{MITCBCL} consists of images of 10 participants captured under varying poses and illumination conditions. The source and the target domains respectively consist of the frontal and the profile view images of the participants, with a total of 360 images in each domain. Some images from the data set are shown in Figure \ref{fig:sample_images_mit}. Matched nodes consist of a pair of images of the same person captured under the same illumination conditions. The images are downsampled to a resolution of $30 \times 30$ pixels. The source and the target graphs are constructed with $K=5$ nearest neighbors and Gaussian scale parameter $\sigma=0.2$.

\textbf{COIL-20 object image data set.} The COIL-20 data set \cite{COIL-20} consists of a total of 1440 images of 20 objects. Each object has 72 images taken from different viewpoints rotating around it. We use this data set as follows for domain adaptation. The 20 objects in the data set are divided into two groups such that each object in the first group forms a pair with the object in the second group that is the most similar to it. The first and the second groups, each of which consists of the images of 10 different objects, are taken as the source domain and the target domain. Two objects that form a pair are considered to have the same class label. The 20 objects in the data set are shown in Figure \ref{fig:COIL20_classes}. Matched nodes consist of images of paired objects captured under the same viewpoint.  The images are downsampled to a resolution of $32 \times 32$ pixels. The graphs are constructed with $K=5$ nearest neighbors and Gaussian scale parameter $\sigma=0.2$. 

 \textbf{Multilingual text data set.}  The multilingual text data set \cite{Amini} contains 6 classes of documents written originally in one language and translated into other languages. Documents are represented with bag-of-words feature vectors obtained using a TFIDF-based weighting scheme \cite{Ramos_usingtf-idf}. The source and the target domains are taken respectively as English and French document sets, each of which contains a total of 1500 documents. A matched node pair consists of an English document and its translation into French. The dimension of feature vectors is reduced to 1000 with PCA as preprocessing. The graphs are constructed with $K=25$ nearest neighbors and edge weights are computed using cosine similarity.

 \textbf{Social network data set.} The Facebook data set \cite{leskovec2012learning} consists of several subnetworks extracted from the Facebook network.  Nodes represent Facebook users and  edges indicate the friendship relationship between users in the graph of each network. The two networks representing different user communities with $168$ and $61$ users illustrated in Figure \ref{fig:Facebook_dataset} are taken as the source and the target graphs. The two graphs contain 27 common users, which are deployed as matched nodes in our experiments. The data set contains information about users such as their education, work, or political affiliations. We have chosen the ``gender" information as the label to be predicted, since it is provided for almost all users. The edge weight is taken as $1$ if a friendship relation exists between a pair of users.



	\begin{figure}[t]
		\centering
		\begin{subfigure}[t]{0.45\textwidth}
			\centering
		   \includegraphics[width=4cm]{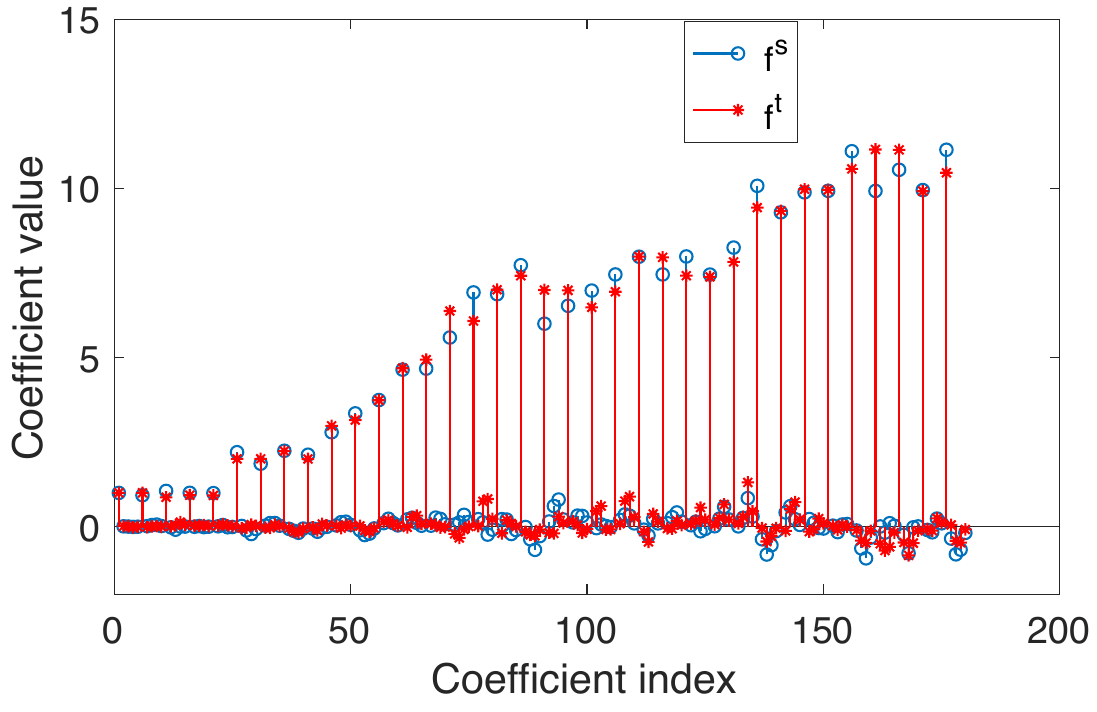}
		\caption{MIT-CBCL data set}
		\label{fig:wavelet_coef_mit}
		\end{subfigure}%
				~
		\begin{subfigure}[t]{0.45\textwidth}
			\centering
		   \includegraphics[width=4cm]{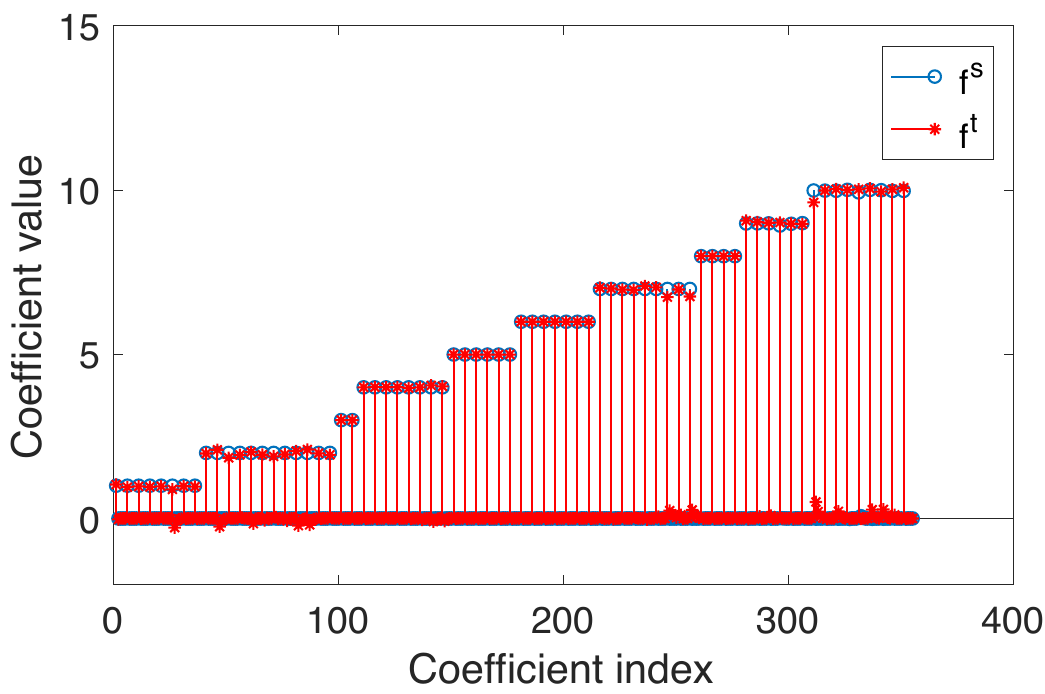}
		\caption{COIL-20 data set}
		\label{fig:wavelet_coef_coil}
		\end{subfigure}%
		\\	
		\begin{subfigure}[t]{0.45\textwidth}
			\centering
		   \includegraphics[width=4cm]{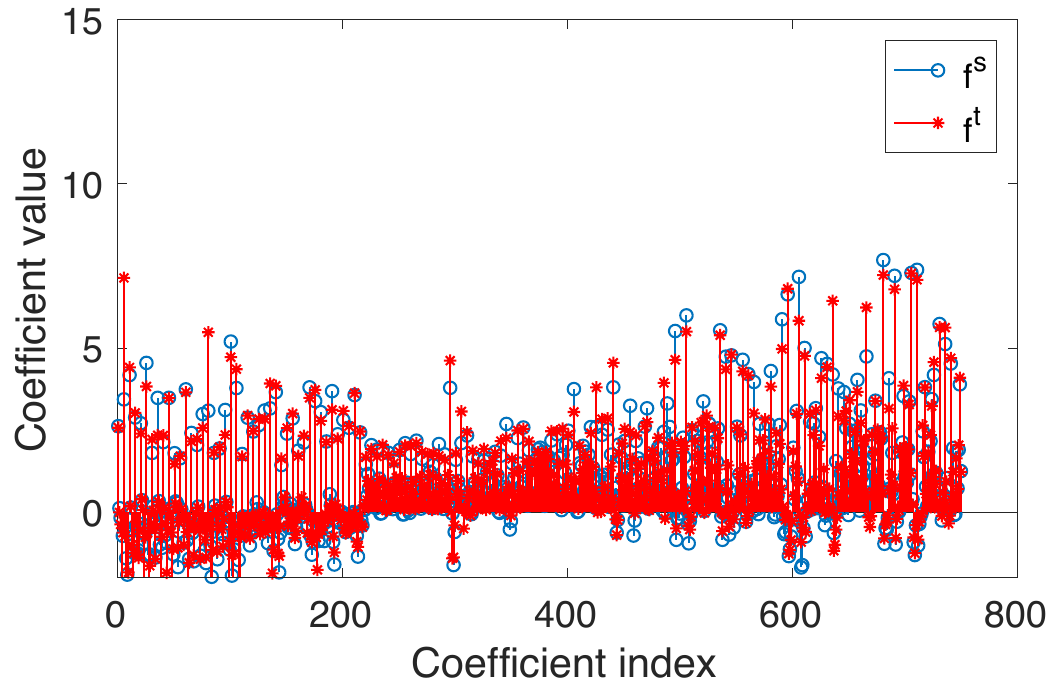}
		\caption{Multilingual data set}
		\label{fig:wavelet_coef_multilingual}
		\end{subfigure}%
		~
		\begin{subfigure}[t]{0.45\textwidth}
			\centering
		   \includegraphics[width=4cm]{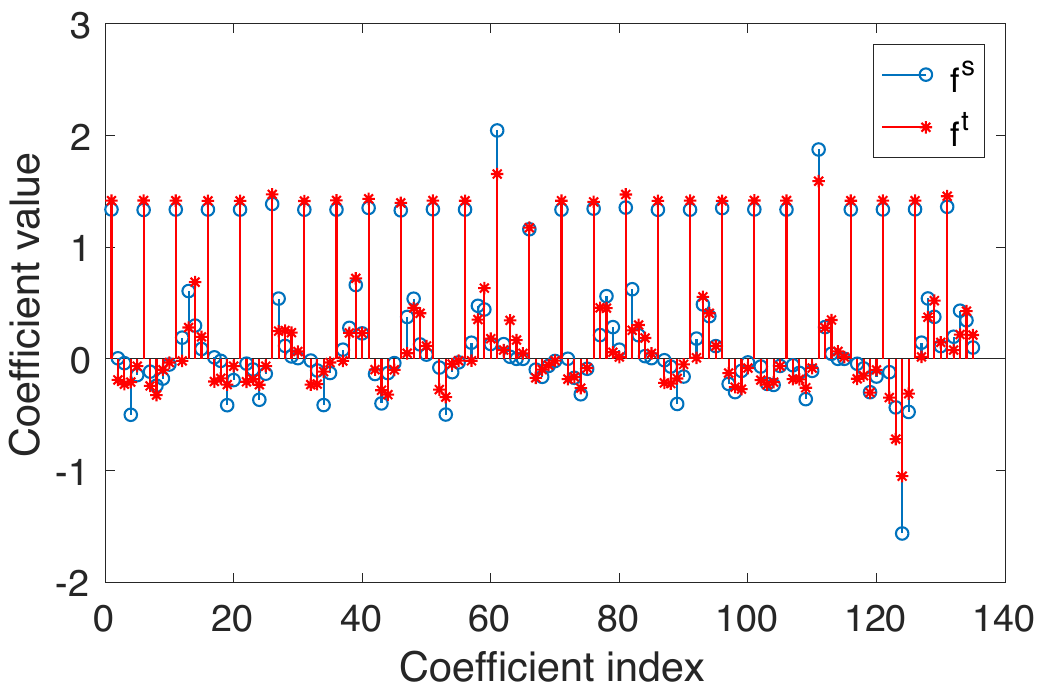}
		\caption{Facebook data set}
		\label{fig:wavelet_coef_facebook}
		\end{subfigure}%
		\caption{Projection coefficients of the label functions $f^s$ and $f^t$ onto the scaling function and wavelets on the graphs. The coefficient indices follow the order of the matched node pair indices and indicate the projections of $f^s$ onto $\phi^s_{m_i}, \psi^s_{s_1, m_i}, \psi^s_{s_2, m_i}, \psi^s_{s_3, m_i}, \psi^s_{s_4, m_i}$  on the source graph and similarly for $f^t $ on the target graph, at each matched node pair $(x^s_{m_i}, x^t_{n_i}) $.}
		\label{fig:coef_match_assump}
	\end{figure}

\subsection{Comparison of the wavelet coefficients between the source and target graphs} 
\label{ssec:comp_wavelet_coef}

We first verify the assumption that the wavelet coefficients of the source and the target label functions have similar values on the two graphs, which is central to the proposed method. The projection coefficients of the label functions $f^s$ and $f^t$ onto the scaling and the wavelet functions on the source and the target graphs are plotted in Figure \ref{fig:coef_match_assump} for the four data sets.  The scaling functions and the wavelets are localized around 36, 71, 150,  and 27 different matched node pairs, respectively for the MIT-CBCL, COIL-20, Multilingual and the Facebook data sets in Figure \ref{fig:coef_match_assump}. The wavelet coefficients plotted in Figure \ref{fig:coef_match_assump} support our assumption and suggest that the projection coefficients at the matched node pairs indeed bear high similarity between the source and the target graphs, despite the fact that the two graphs are constructed independently. This is also verified quantitatively by computing the following normalized coefficient dissimilarity measure 
\[
\Delta=\frac{\sum_i (c^s_i  - c^t_i)^2 }{\sum_i(c^s_i)^2}
\]
for all data sets, where $c^s_i$ and $c^t_i$ simply denote the $i$-th projection coefficient on the source and the target graphs. The normalized dissimilarity $\Delta$ is found to have the considerably small values 0.0119, 0.0005, 0.0711, and 0.0521, respectively for the MIT-CBCL, COIL-20, Multilingual and the Facebook data sets; thus confirming the hypothesis that the source and the target wavelet coefficients take similar values in typical domain adaptation setups.

\subsection{Comparative Experiments}

The proposed GrALP algorithm is compared to the domain adaptation methods  Subspace Alignment (SA) \cite{Fernando2013}, Easy Adapt++ (EA++) \cite{Daume2010}, Geodesic Flow Kernel (GFK) \cite{GongSSG12}, Joint Geometrical and Statistical Alignment (JGSA) \cite{ZhangLO17}, Scatter Component Analysis (SCA) \cite{GhifaryBKZ17}, LDA-Inspired Domain Adaptation (LDADA) \cite{LuSC0H18}, Domain Adaptation via Spectral Graph Alignment (DASGA) \cite{PilanciV20}; in addition to the basic classifiers Support Vector Machine (SVM), Nearest Neighbor classification (NN), and the graph-based Semi-Supervised Learning with Gaussian fields (SSL) method  \cite{ZhuGL03}. A comparison of the properties of the domain adaptation methods used in the experiments is given in Table \ref{tab:compare_alg_prop}.

\begin{table}[]
\footnotesize
\centering
\caption{Comparison of the properties of the domain adaptation methods used in the experiments. Columns respectively  indicate whether each method uses source labels, target labels, graph models, projections or transformations of feature spaces, and feature augmentation techniques when learning representations.}
\label{tab:compare_alg_prop}
\begin{tabular}{|c|c|c|c|c|c|}
\hline
Algorithm & Source labels & Target labels & Graph model & Feature proj. & Feature aug. \\ \hline
\textbf{SA} \cite{Fernando2013} & \texttimes &   \texttimes & \texttimes  & \checkmark & \texttimes  \\ \hline
\textbf{EA++} \cite{Daume2010}& \checkmark &   \checkmark & \texttimes  & \texttimes & \checkmark  \\ \hline
\textbf{GFK} \cite{GongSSG12} & \texttimes &   \texttimes & \texttimes  & \checkmark & \texttimes  \\ \hline
\textbf{JGSA} \cite{ZhangLO17} & \checkmark &   \texttimes & \texttimes  & \checkmark & \texttimes  \\ \hline
\textbf{SCA} \cite{GhifaryBKZ17} & \checkmark &   \checkmark & \texttimes  & \checkmark & \texttimes  \\ \hline
\textbf{LDADA} \cite{LuSC0H18} & \checkmark &   \texttimes & \texttimes  & \checkmark & \texttimes  \\ \hline
\textbf{DASGA} \cite{PilanciV20} & \checkmark &   \checkmark & \checkmark  & \texttimes &  \texttimes \\ \hline
\textbf{GrALP} & \checkmark &   \checkmark & \checkmark  & \texttimes & \texttimes  \\ \hline
\end{tabular}
\end{table}

In all experiments, misclassification rates of the methods over the unlabeled target samples are evaluated in the following settings: 

\textbf{Target sweep:} The matched node pairs are unlabeled and all other source samples are labeled. The ratio of the known target labels is varied.

\textbf{Source sweep:} The matches are unlabeled and no label information is available in the target domain. The ratio of known source labels is varied. 

\textbf{Unlabeled match sweep:} The matches are unlabeled and no label information is available in the target domain. The ratio of matched nodes is varied.

\textbf{Partially labeled match sweep:} A certain percentage of the matches and the source samples are labeled. The unmatched target samples are unlabeled. The ratio of matched nodes is varied. 

\textbf{Labeled match sweep:} All matches and a certain percentage of the source samples are labeled. The unmatched target samples are unlabeled. The ratio of matched nodes is varied.


All methods are provided with all label information that leaks between the two domains through the matches. The SVM and the NN methods are trained over all labeled samples in the source and the target domains and the SSL method is applied on the target graph, which are the settings yielding the best performance. The SCA, DASGA, and SSL algorithms require labeled samples in the target domain, therefore, are excluded from the experiments with no labeled target samples. For the GrALP method, the algorithm parameters are selected as $\mu=1$, ${\gamma}_s=0.1$, and ${\gamma}_t=0.1$ in all experiments and one-hot label vectors are used for representing the label functions. The parameters of the other algorithms are optimized for each data set in order to attain the best performance. The parameter settings of the algorithms that require parameter tuning are reported in Table \ref{tab:alg_param} for all data sets. The results obtained on different datasets are presented below.

\begin{table}[]
\footnotesize
\centering
\caption{Parameters of the algorithms used in experiments (denoted as in the original papers)}
\label{tab:alg_param}
\begin{tabular}{|c|c|c|c|c|c|c|c|}
\hline
Algorithm& Parameter & MIT-CBCL & COIL-20 & Multilingual & Facebook \\ \hline
\textbf{SA}  \cite{Fernando2013} & $d$ &   37 & 57  & 1000 & 5  \\ \hline
 &  $k$ & 23 & 38 & 30 & 3  \\
 & $T$ & 19 & 15 & 20 & 20 \\ 
 \textbf{JGSA} \cite{ZhangLO17} & $\lambda$  & 2.98 & 0.83 & 0.1 & 1   \\ 
 & $\mu$  & 1 & 0.065 & 100 & 100   \\
 & $\beta$  & 0.01 & 6 & 0.1 & 1   \\ \hline
\textbf{LDADA} \cite{LuSC0H18}&  $T$ & 5 & 10 & 10 & 10  \\ \hline
 &  $\beta$ & 0.01 & 0.02 & 0.0001 & 1  \\
  \textbf{SCA} \cite{GhifaryBKZ17}  & $\delta$ & 0.01 & 0.01 & 10 & 0.1 \\ 
 & $k$  & 9 & 9 & 5 & 5   \\ \hline
 &  $\mu_1$ & 0.01 & 1 & 0.1 & 1  \\
 \textbf{DASGA} \cite{PilanciV20}  & $\mu_2$ & 1 & 1 & 1 & 0.1 \\ 
  & $R$  & 9 & 10 & 6& 6   \\ \hline
\end{tabular}
\end{table}

	\begin{figure}[t]
		\centering
		\begin{subfigure}[t]{0.45\textwidth}
			\centering
			\includegraphics[height=4cm]{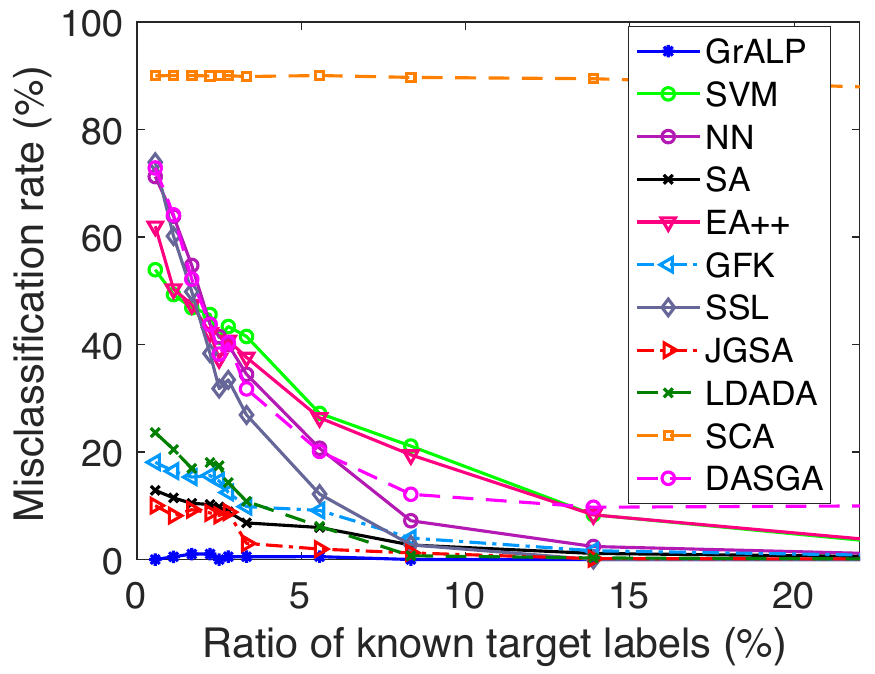}
			\caption{Target sweep}
		\label{fig:errors_face_target_sweep}
		\end{subfigure}
		~
		\begin{subfigure}[t]{0.45\textwidth}
			\centering
		   \includegraphics[height=4cm]{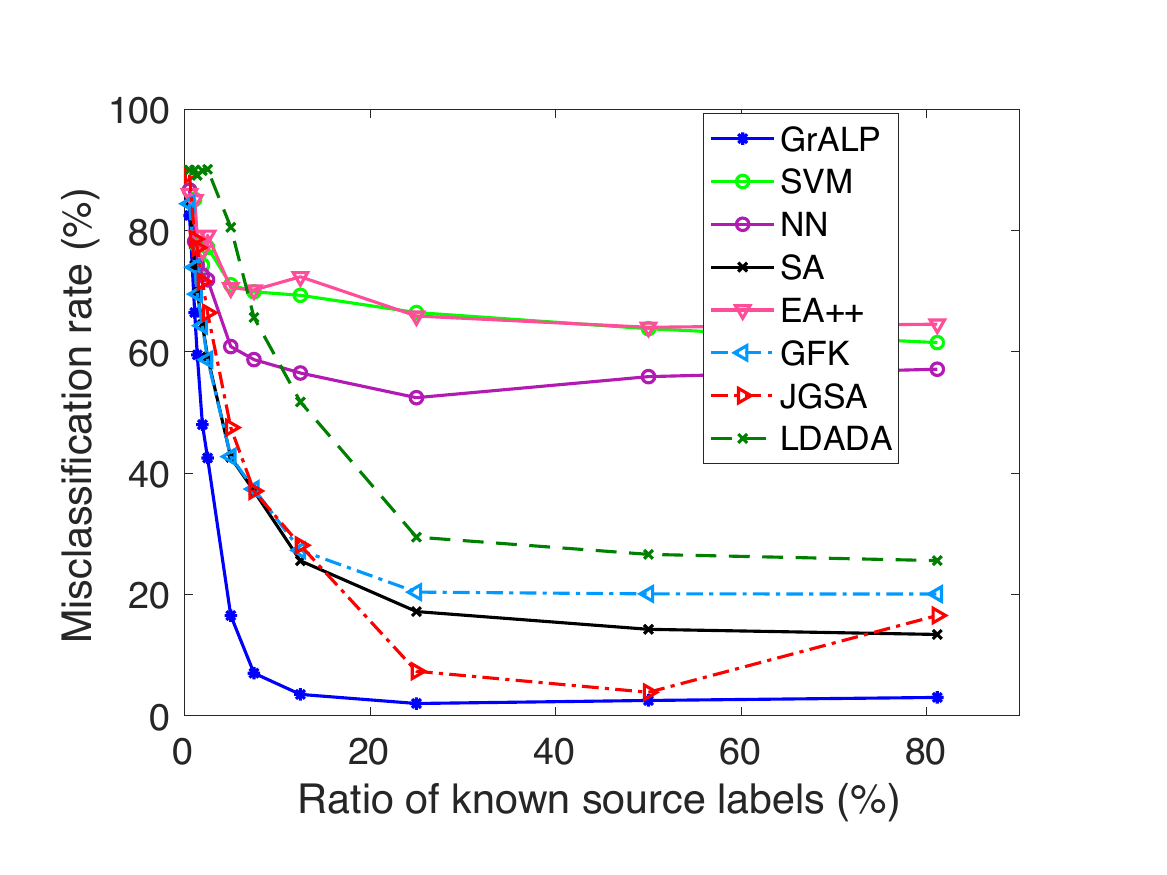}
		\caption{Source sweep}
		\label{fig:errors_face_source_sweep}
		\end{subfigure}\\%
		~
				\begin{subfigure}[t]{0.31\textwidth}
			\centering
			\includegraphics[height=3cm]{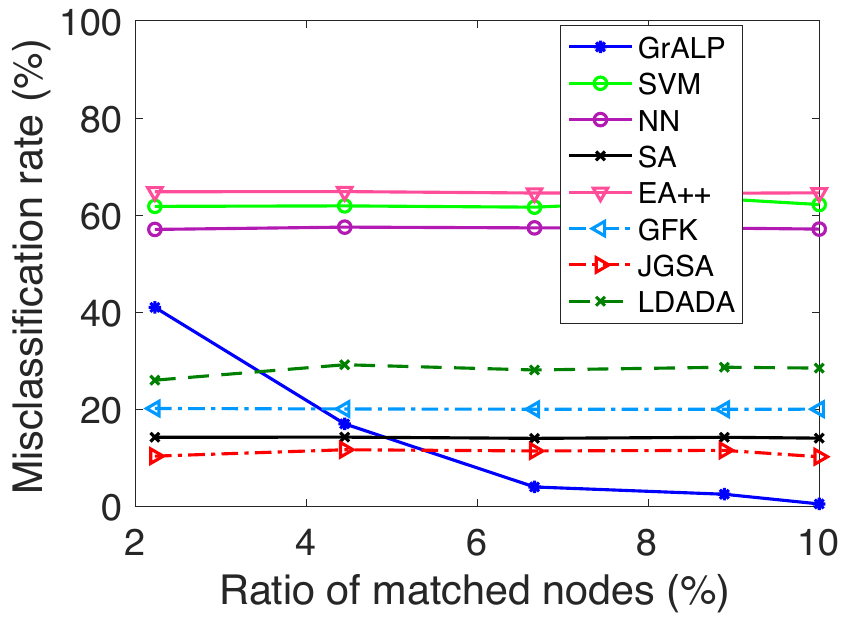}
			\caption{Unlabeled match sweep}
		\label{fig:errors_face_match_sweep_324}
		\end{subfigure}
	~
				\begin{subfigure}[t]{0.31\textwidth}
				\centering
		\includegraphics[height=3cm]{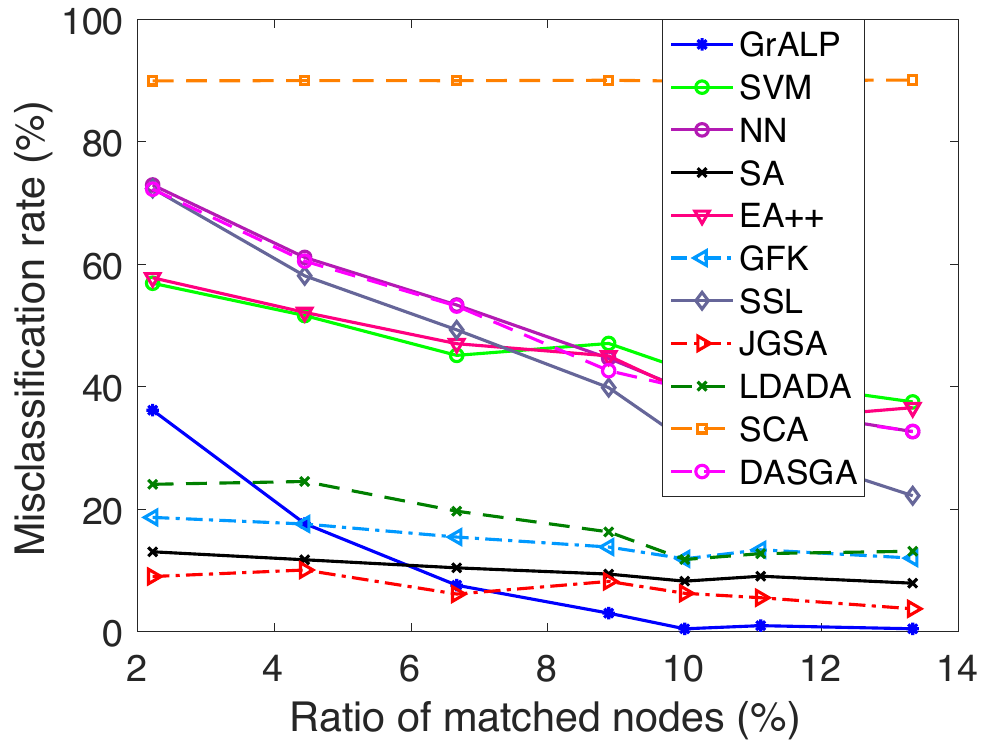}
		\caption{P. labeled match sweep}
		\label{fig:errors_face_match_part_labeled}
	\end{subfigure}
		~
		 	\begin{subfigure}[t]{0.31\textwidth}
				\centering
		\includegraphics[height=3cm]{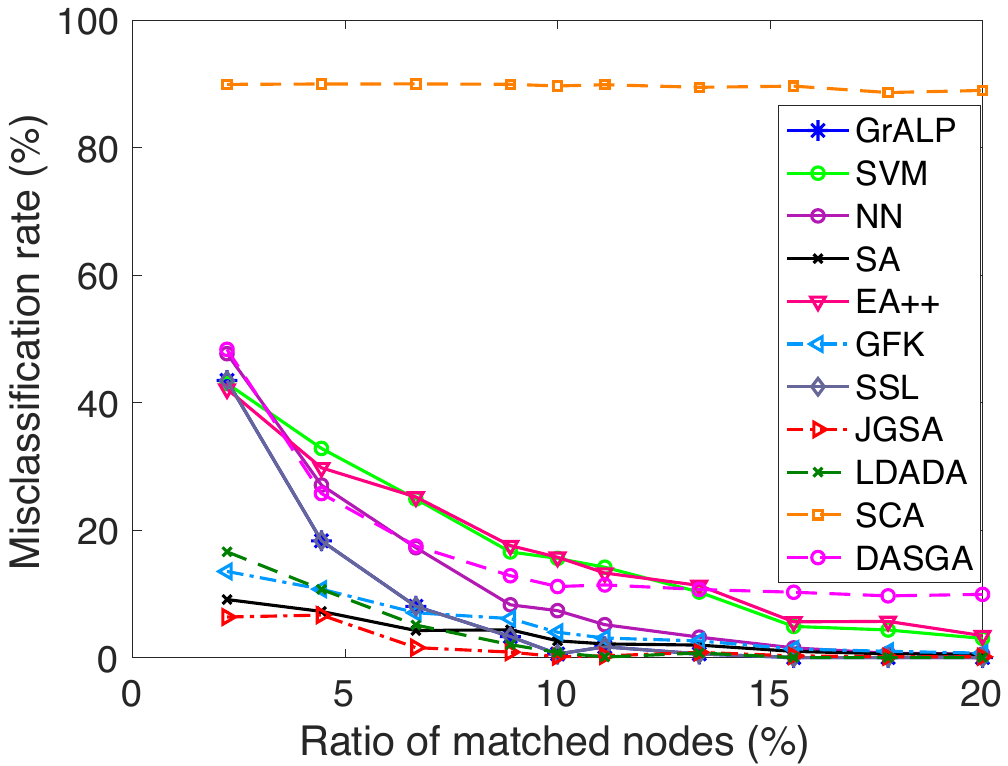}
		\caption{Labeled match sweep}
		\label{fig:errors_face_match_all_labeled}
	\end{subfigure}
		\caption{Target misclassication rates for the MIT-CBCL data set.  In panels (a) and (b) $10\%$ of the nodes are matched. In panels (c) and (e), $90\%$ of the source labels are known. In panel (d), $25\%$ of the matches and $90\%$ of the source samples are labeled.}\label{fig:MIT-CBCL}
	\end{figure}
	

	
For the MIT-CBCL data set, the misclassification rate of the unlabeled target samples is plotted in percentage for all methods in Figure \ref{fig:MIT-CBCL}. All results are averaged over 20 repetitions with random selections of matched samples and labeled samples. The proposed GrALP method gives the best performance in Figures \ref{fig:errors_face_target_sweep} and \ref{fig:errors_face_source_sweep}. The domain adaptation methods GrALP, LDADA, JGSA, GFK, and SA yield much higher classification accuracy than the other algorithms when a small set of target labels are available. As expected, domain adaptation algorithms perform better than basic classification methods. In the target sweep setting in Figure \ref{fig:errors_face_target_sweep}, the proposed GrALP method is observed to provide almost zero classification error, even for a very small percentage of target labels. Similarly, in the source sweep setting in Figure \ref{fig:errors_face_source_sweep}, the target error of GrALP drops quite quickly as the ratio of labeled source nodes starts increasing. GrALP can use source labels more effectively than the other methods as it employs the information of the matched node pairs. Considering that no label information is available in the target domain and the matches are also unlabeled in Figure \ref{fig:errors_face_source_sweep}, we conclude that the proposed GrALP algorithm is able to successfully transfer the label information from the source graph to the target graph through the wavelet coefficients over the matches. The misclassification rate in the target domain is observed to decrease down to $3\%$ although there is no label information in the target domain.

In the unlabeled match sweep setting in Figure \ref{fig:errors_face_match_sweep_324}, no labels are available in the target domain or on the matched nodes. Since the algorithms other than GrALP do not use the information of the matched nodes, their target misclassification rate is not affected by the number of matched nodes. As the ratio of matched nodes increases, the misclassification rate of the proposed GrALP algorithm decreases as expected, reaching zero misclassification rate when $10\%$ of the nodes are matched, despite the strict unavailability of labels in the target domain. While GrALP outperforms the other methods when there is a sufficient number of matches, we observe that the other methods perform better when the number of matches is too small. This is for the following reason: GrALP is a purely graph-based method that does not at all employ the ambient space representations of data samples once the source and the target graphs are constructed. Data samples are simply represented as abstract graph nodes and the only way the algorithm can link the source and the target domains is through the matched nodes. On the other hand, all the other methods (except for SSL) heavily rely on ambient space representations (feature vectors) of data samples. Having access to the physical coordinates of data unlike GrALP, they outperform GrALP when the number of matches is too few. The results of the partially labeled match sweep experiment in Figure \ref{fig:errors_face_match_part_labeled} similarly show that the proposed method outperforms the others, provided that a sufficient number of matches (around $7-8 \%$) is available. In Figure \ref{fig:errors_face_match_all_labeled} where all matches are labeled, the error rate of GrALP becomes very close to 0 when around $10\%$ of the nodes are matched. However, the error rates of some other algorithms also drop near 0 as the number of matches increases, since their access to the label information increases rapidly as all matches are labeled. \\

	\begin{figure}[t]
		\centering
	\begin{subfigure}[t]{0.45\textwidth}
		\centering
		\includegraphics[height=4cm]{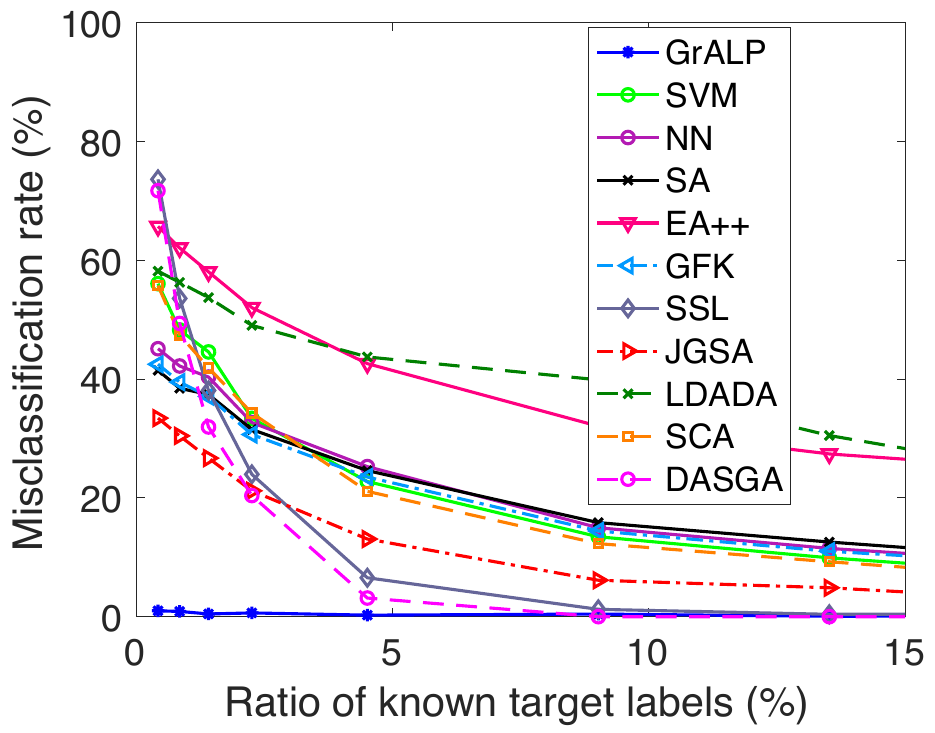}
		\caption{Target sweep}
	     \label{fig:errors_COIL20_target_sweep}
	\end{subfigure}
~
\begin{subfigure}[t]{0.45\textwidth}
		\centering
		{\includegraphics[height=4cm]{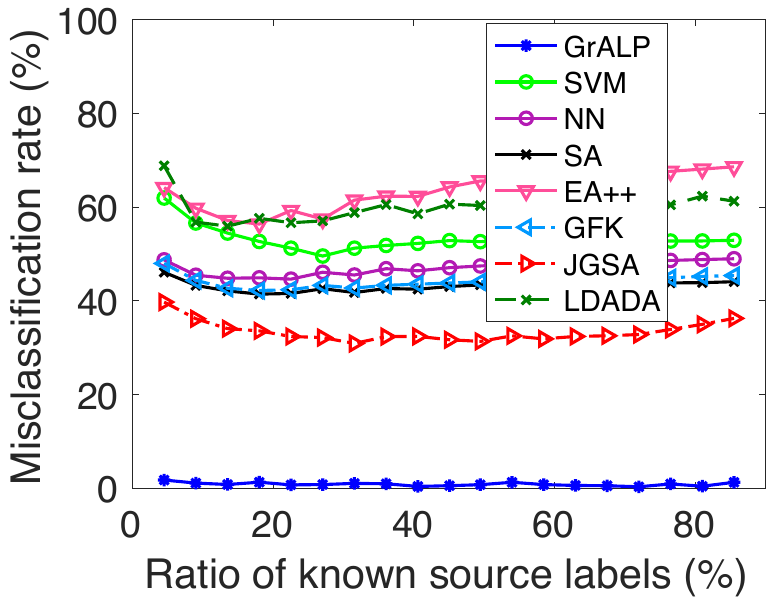}}
		\caption{Source sweep}
	\label{fig:errors_COIL20_source_71_matched_nodes}
	\end{subfigure}\\
~
		\begin{subfigure}[t]{0.31\textwidth}
			\centering
			\includegraphics[height=3cm]{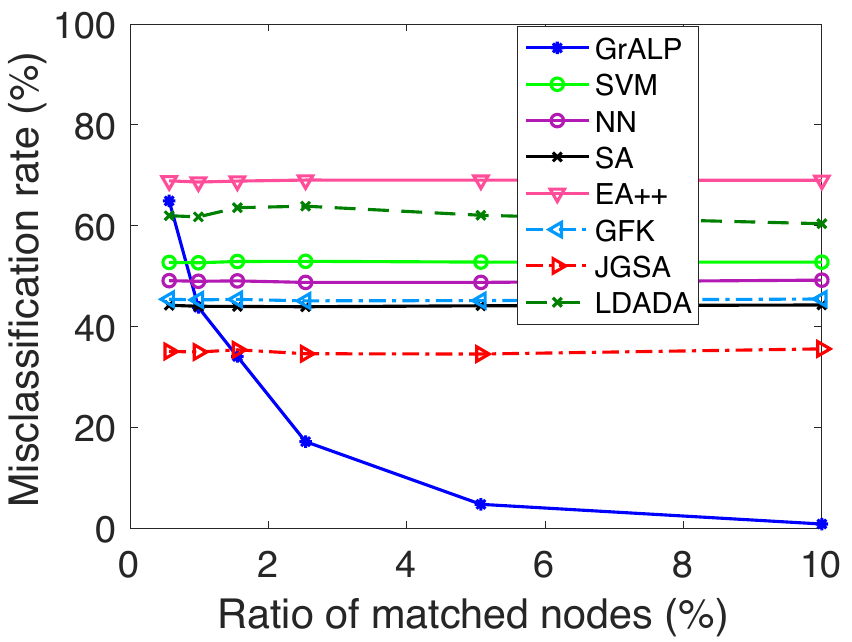}
			\caption{Unlabeled match sweep}
			\label{fig:errors_COIL20_unlabeled_match_639}
		\end{subfigure}
		~
			\begin{subfigure}[t]{0.31\textwidth}
		\centering
		\includegraphics[height=3cm]{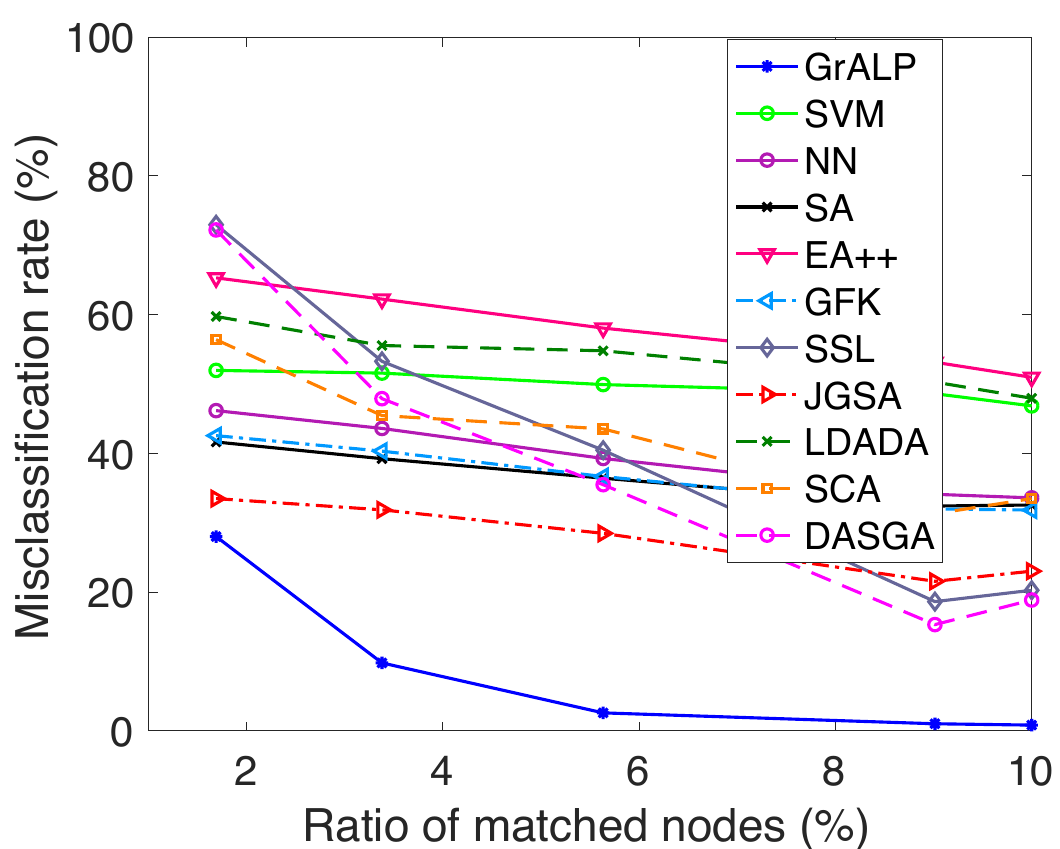}
		\caption{P. labeled match sweep}
		\label{fig:errors_COIL20_match_part_labeled}
			\end{subfigure}	
				~	
		\begin{subfigure}[t]{0.31\textwidth}
		\centering
		\includegraphics[height=3cm]{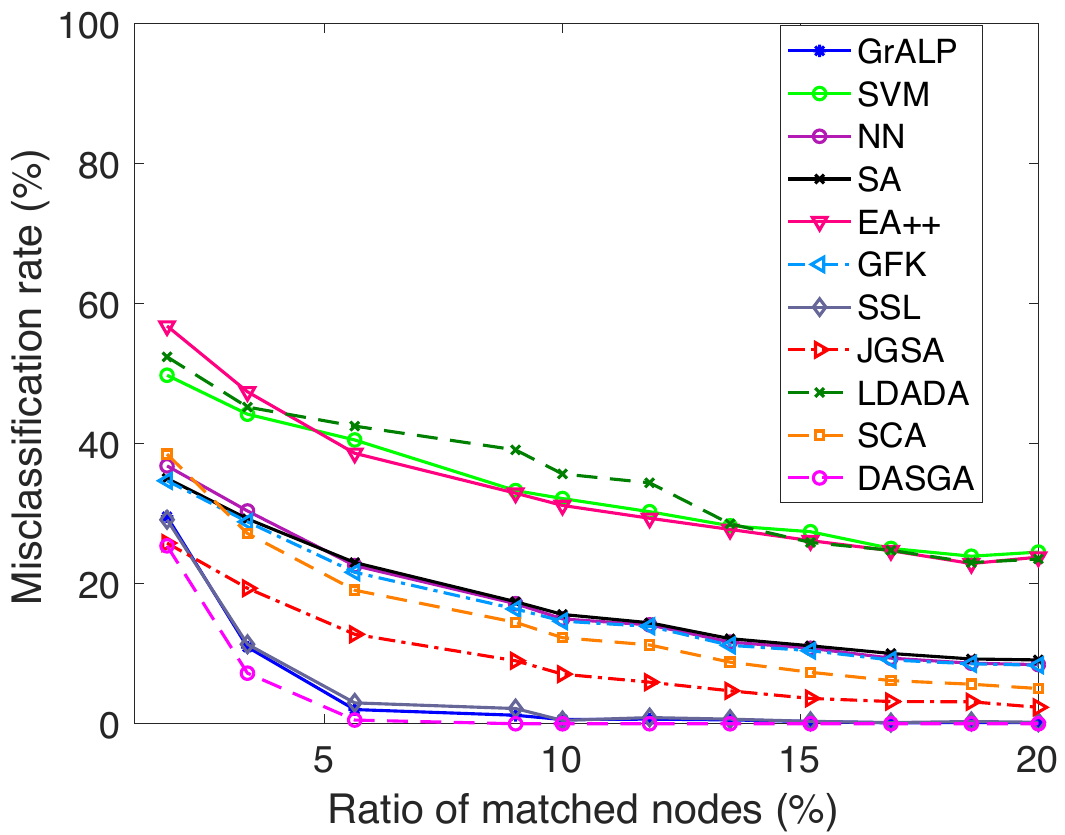}
		\caption{Labeled match sweep}
		\label{fig:errors_COIL20_match_all_labeled}
			\end{subfigure}				
		\caption{Target misclassication rates for the COIL-20 data set.  In panels (a) and (b) $10\%$ of the nodes are matched. In panels (c) and (e), $90\%$ of the source labels are known. In panel (d), $25\%$ of the matches and $90\%$ of the source samples are labeled.}\label{fig:COIL-match}
\end{figure}

	



 The results obtained on the COIL-20 data set are presented in Figure \ref{fig:COIL-match}, which are averaged over $20$ random repetitions of the experiment. In Figures \ref{fig:errors_COIL20_target_sweep} and \ref{fig:errors_COIL20_source_71_matched_nodes} the proposed GrALP method outperforms the other methods and yields quite high classification accuracy even for a very small number of labeled nodes. In Figure \ref{fig:errors_COIL20_unlabeled_match_639} the classification accuracy of GrALP exceeds that of the other methods as soon as the ratio of matched nodes attains $2-3 \%$ when the matches are unlabeled, while its performance is seen to be even better in Figure \ref{fig:errors_COIL20_match_part_labeled} in case of partially labeled matches. The proposed method performs particularly well in this data set. Data samples are regularly sampled from the data manifold, resulting in an even and regular graph structure. This contributes positively to the accuracy of graph-based methods. One can indeed observe that, being purely graph-based methods, SSL and DASGA also achieve high classification accuracy in Figures \ref{fig:errors_COIL20_target_sweep} and \ref{fig:errors_COIL20_match_all_labeled}. In Figure \ref{fig:errors_COIL20_target_sweep} GrALP outperforms SSL and DASGA as few labels are known and GrALP exploits the information transferred from the source graph through the local wavelet coefficients; while in Figure \ref{fig:errors_COIL20_match_all_labeled} SSL and DASGA attain the performance of GrALP, owing to the fact that more labels are known in the target graph as all matches are labeled.  \\

\begin{figure}[t]
		\centering
		\begin{subfigure}[b]{0.4\textwidth}
			\centering
		   \includegraphics[height=4cm]{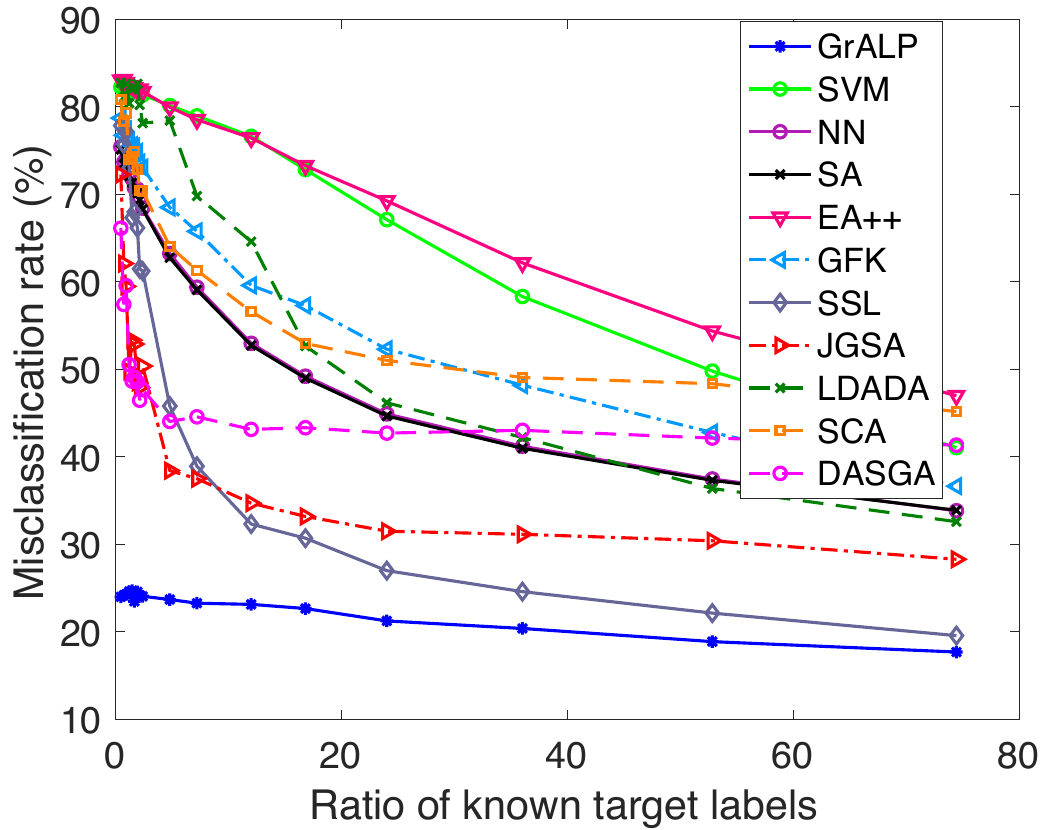}
		\caption{Target sweep}
		\label{fig:errors_multilingual_target}
		\end{subfigure}%
		~ 
		\begin{subfigure}[b]{0.4\textwidth}
			\centering
			\includegraphics[height=4cm]{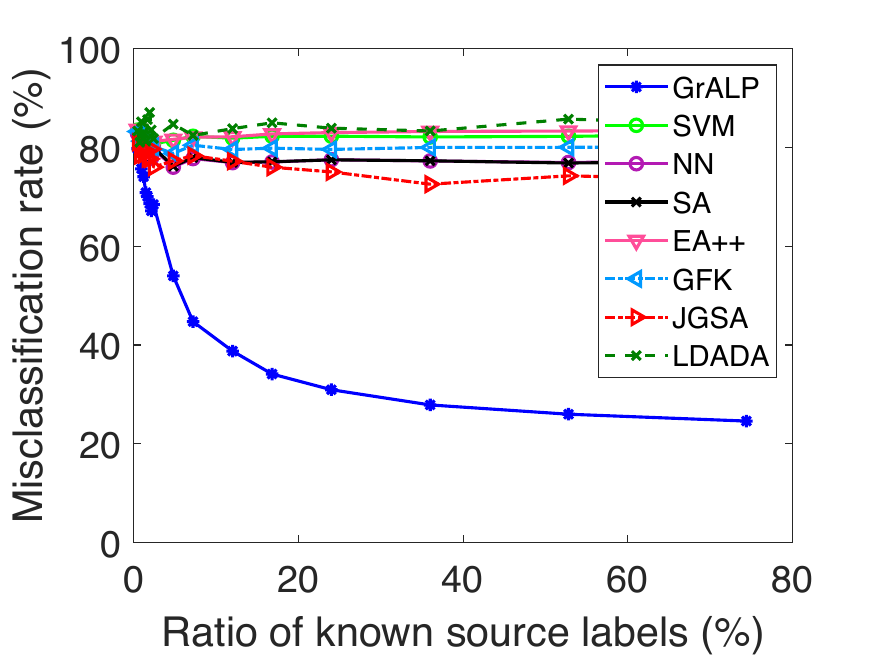}
			\caption{Source sweep}
			\label{fig:errors_multilingual_source}
		\end{subfigure}
		\\
		\begin{subfigure}[b]{0.31\textwidth}
			\centering
			\includegraphics[height=3cm]{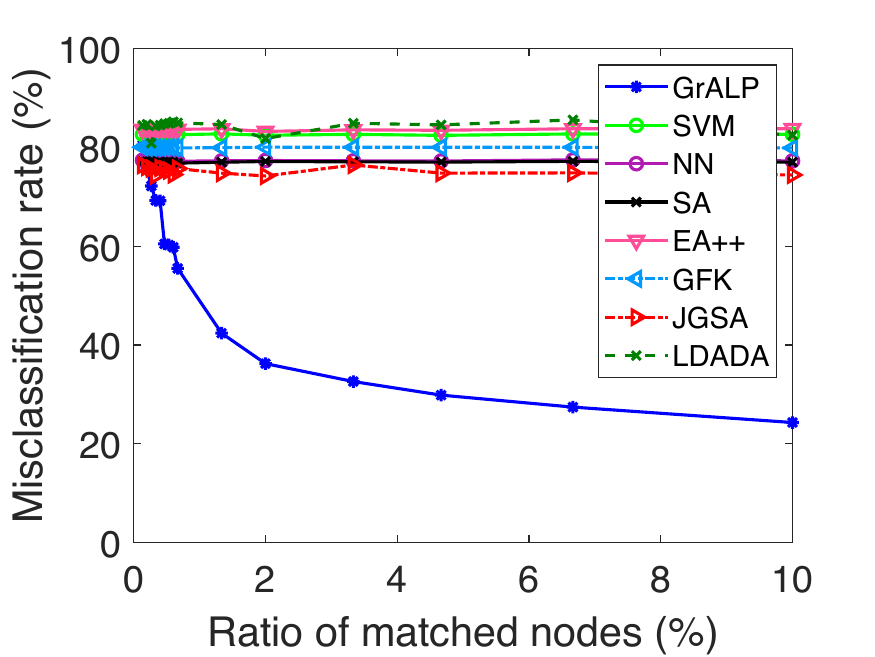}
			\caption{Unlabeled match sweep}
			\label{fig:errors_multilingual_match}
		\end{subfigure}
			~ 
				\begin{subfigure}[b]{0.31\textwidth}
			\centering
			\includegraphics[height=3cm]{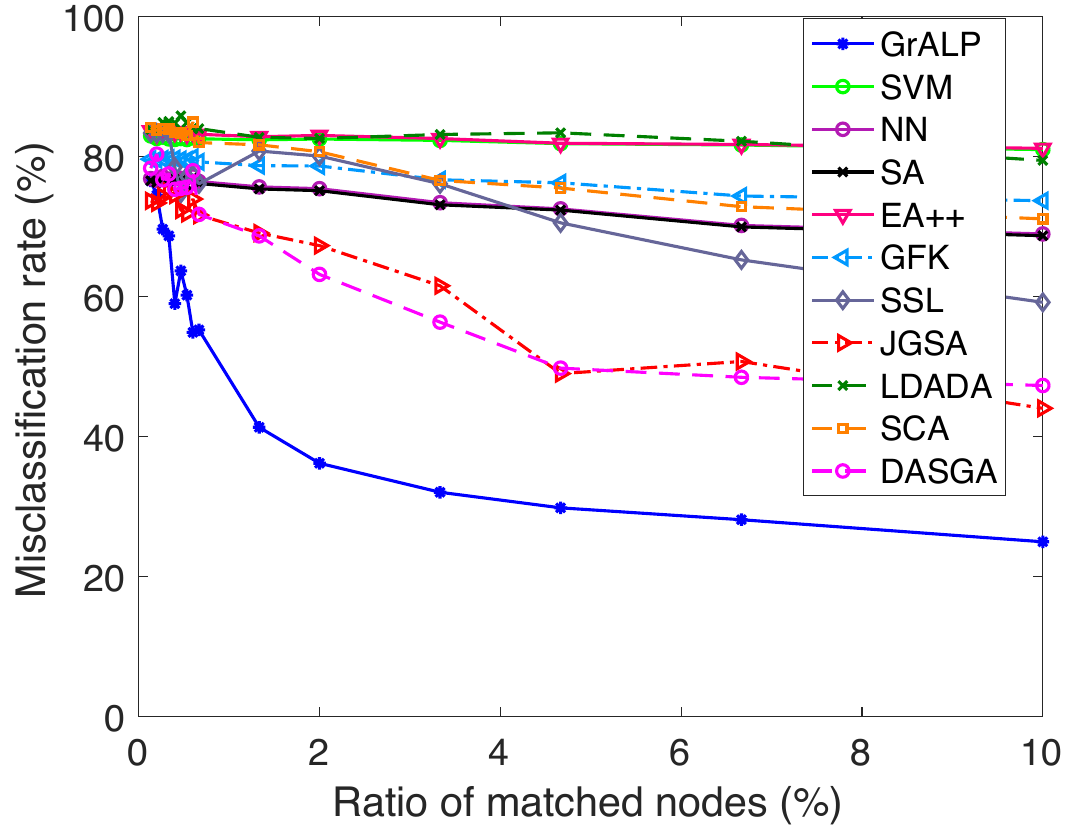}
			\caption{P. labeled match sweep}
			\label{fig:errors_multilingual_part_labeled}
		\end{subfigure}
		~
			\begin{subfigure}[b]{0.31\textwidth}
			\centering
			\includegraphics[height=3cm]{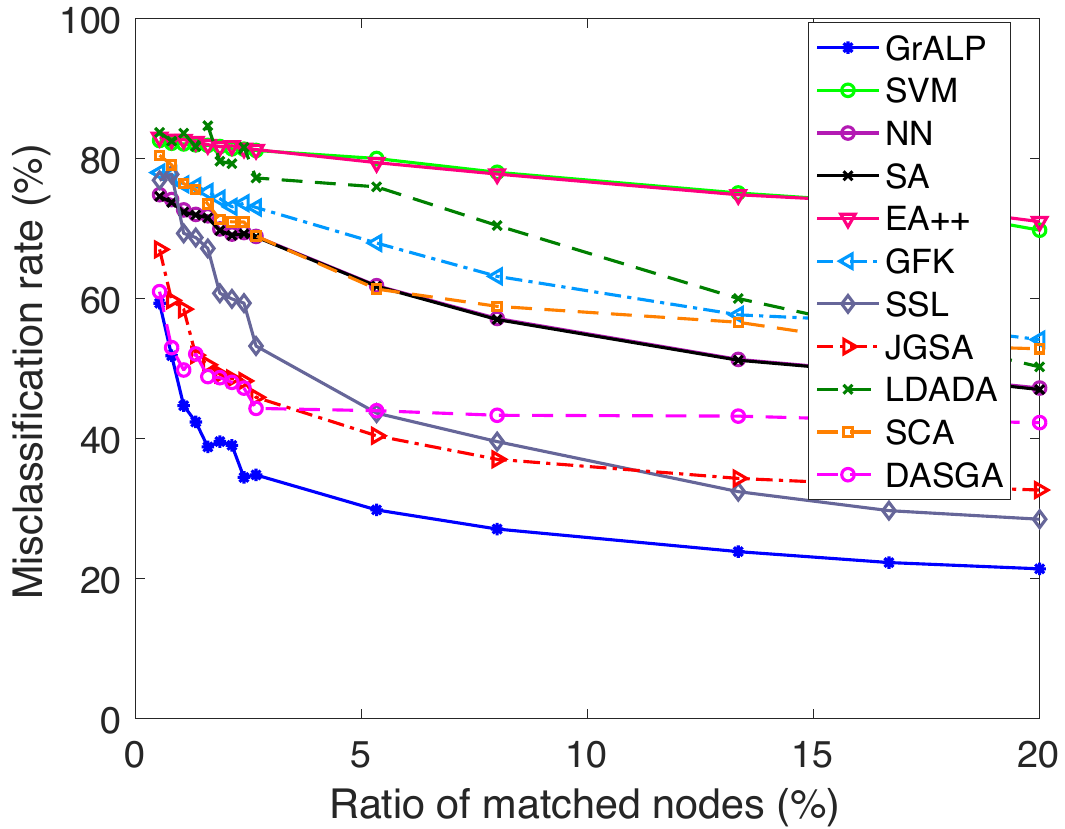}
			\caption{Labeled match sweep}
			\label{fig:errors_multilingual_all_labeled}
		\end{subfigure}
		\caption{Target misclassication rates for the Multilingual text data set.  In panels (a) and (b) $10\%$ of the nodes are matched. In panels (c) and (e), $90\%$ of the source labels are known. In panel (d), $25\%$ of the matches and $90\%$ of the source samples are labeled.}\label{fig:Multilingual}
		\vspace{-0.5cm}
	\end{figure}

 
Next, Figure \ref{fig:Multilingual} shows the results obtained on the Multilingual text data set,  which are averaged over $10$ random repetitions of the experiment. The proposed GrALP method performs quite well in this data set.  Even with a very small amount of matched nodes, the misclassification rate of GrALP is lower than that of the other methods. While the performances of the other domain adaptation methods consistently improve with the increase in the target labels in Figure \ref{fig:errors_multilingual_target}, their performances improve slowly or stagnate with the increase in the source labels or matched nodes in Figures \ref{fig:errors_multilingual_source} and \ref{fig:errors_multilingual_part_labeled}. We interpret this in the way that the bag-of-words feature representations of documents written in different languages are not easy to align by transformations or projections onto a common domain, therefore, the information available in the source domain cannot be not exploited efficiently. On the other hand, the graph-based GrALP algorithm performs better as it relies on relating the label information to the affinities between pairs of data samples in the same source graph and transmitting this information to a target graph of similar topology, instead of learning a classifier based on the ambient space representations of data samples.\\

    \begin{figure}[t]
		\begin{subfigure}[b]{0.4\textwidth}
			\centering
		   \includegraphics[height=4cm]{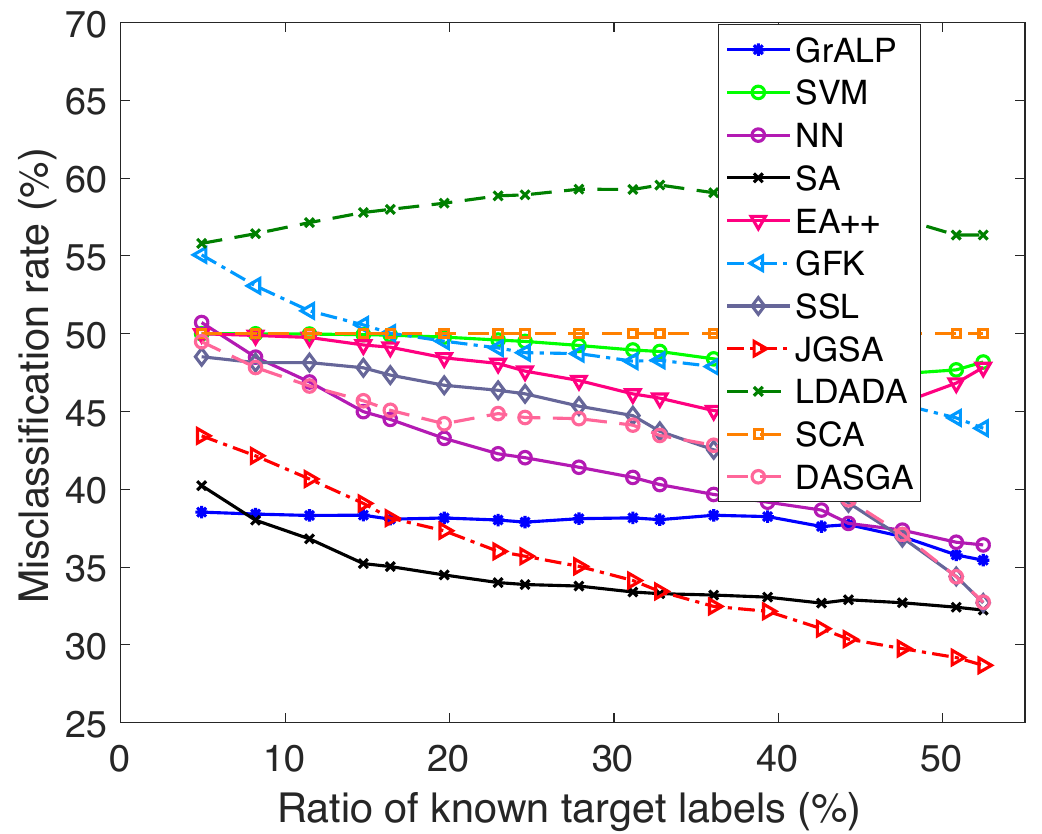}
		\caption{Target sweep}
		\label{fig:errors_facebook_target}
		\end{subfigure}%
		~ 
		\begin{subfigure}[b]{0.4\textwidth}
			\centering
			\includegraphics[height=4cm]{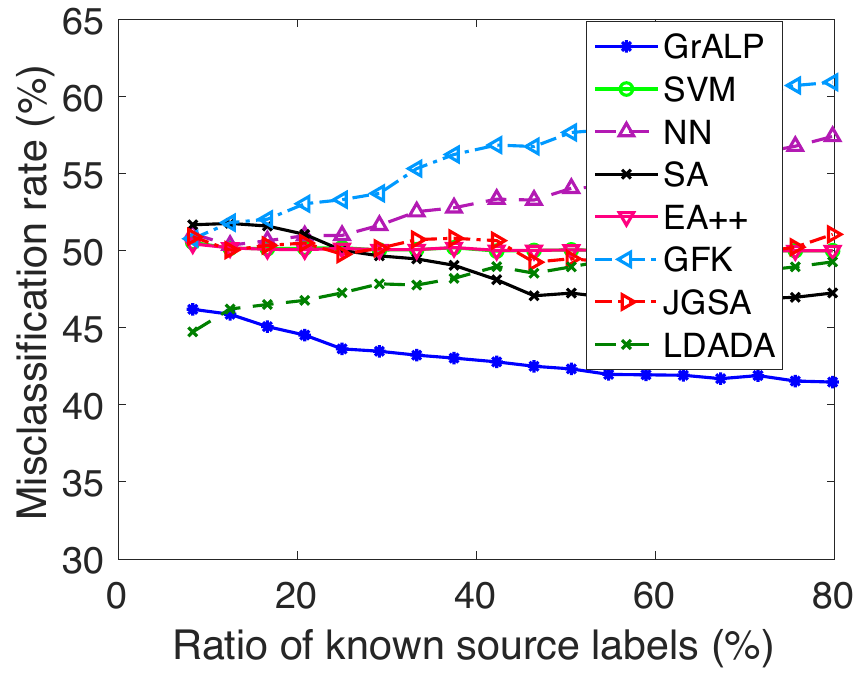}
			\caption{Source sweep}
			\label{fig:errors_facebook_source}
		\end{subfigure}
		\\
		\begin{subfigure}[b]{0.31\textwidth}
			\centering
			\includegraphics[height=3cm]{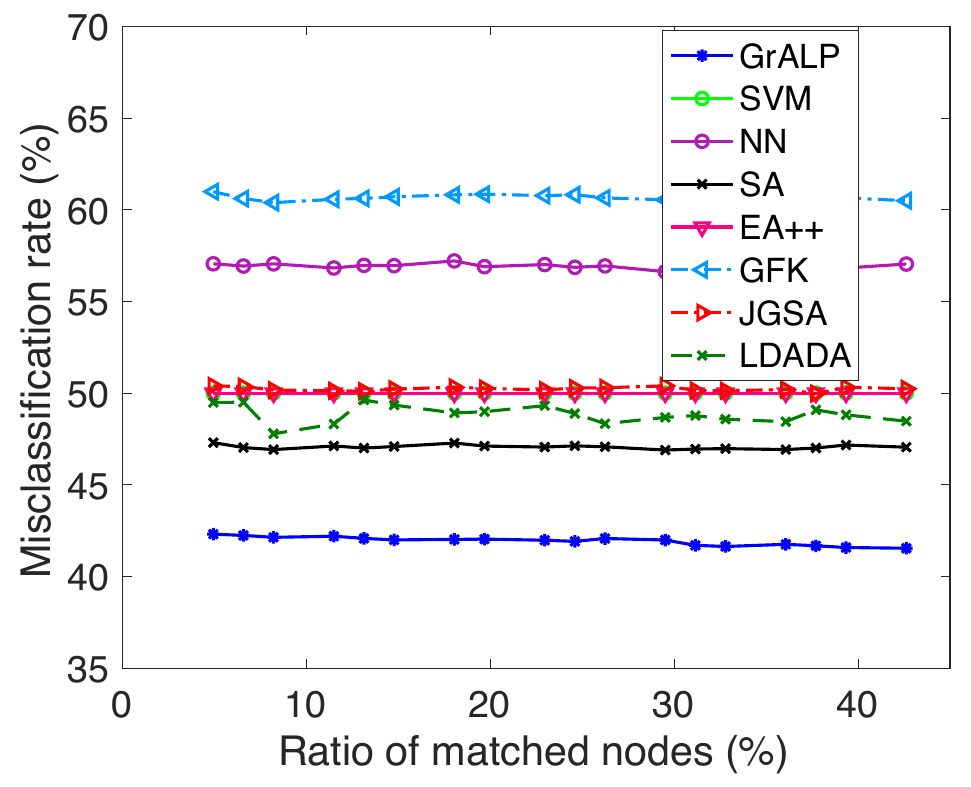}
			\caption{Unlabeled match sweep}
			\label{fig:errors_facebook_match}
		\end{subfigure}
			~ 
				\begin{subfigure}[b]{0.31\textwidth}
			\centering
			\includegraphics[height=3cm]{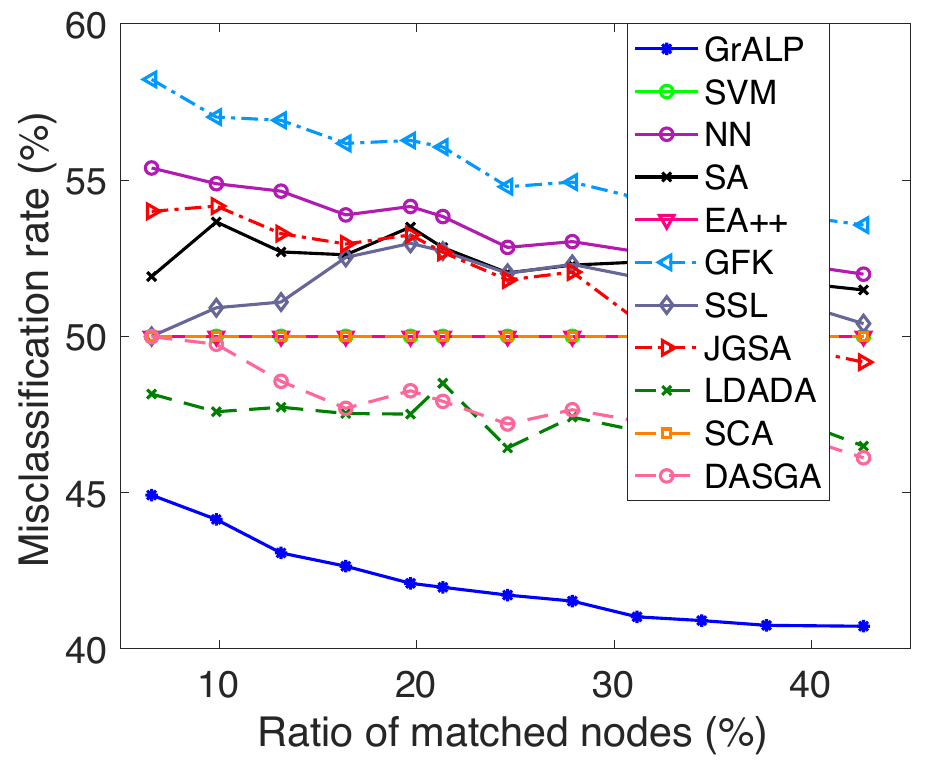}
			\caption{P. labeled match sweep}
			\label{fig:errors_facebook_part_labeled}
		\end{subfigure}
		~
			\begin{subfigure}[b]{0.31\textwidth}
			\centering
			\includegraphics[height=3cm]{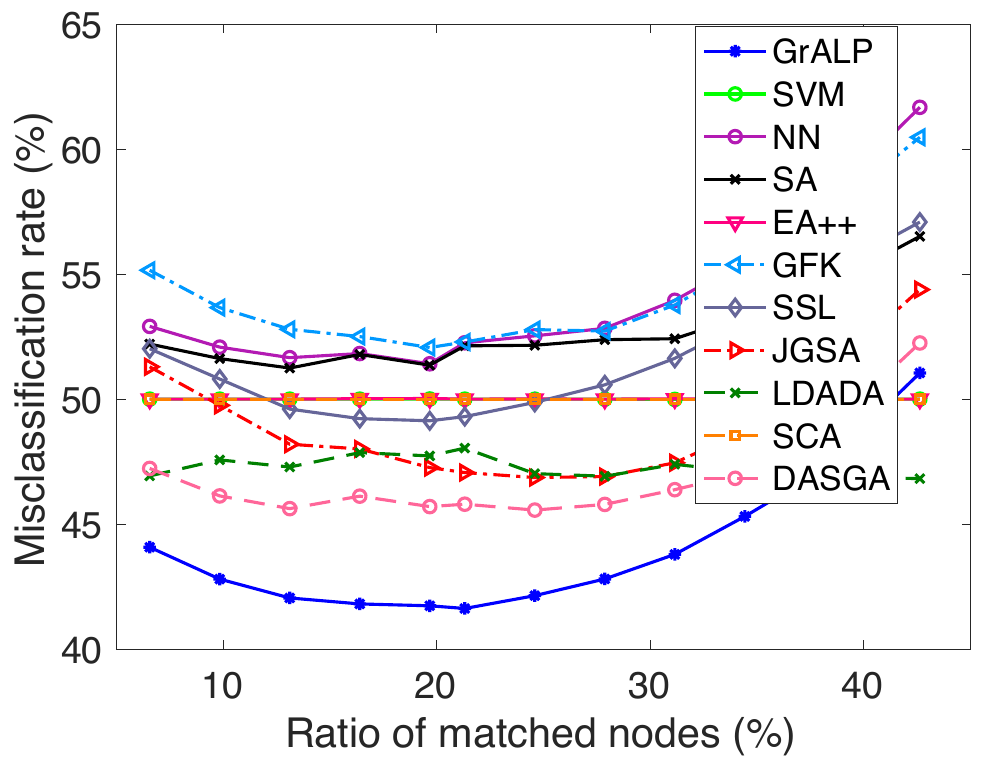}
			\caption{Labeled match sweep}
			\label{fig:errors_facebook_all_labeled}
		\end{subfigure}
		\caption{Target misclassification rates for the Facebook data set. In panels (a) and (b) all 27 common users are used as matched node pairs. In panels (c) and (e), $90\%$ of the unmatched 141 source nodes are labeled. In panel (d),  $25\%$ of the matches are labeled. Balanced misclassification rates are reported.}\label{fig:Facebook}
		\label{fig:resultsFacebook}
	\end{figure}

 Finally, we evaluate the methods on the Facebook data set. Unlike in the previous data sets, data samples are not embedded in an ambient space in this data set, as each data sample represents a social network user. As the methods except GrALP, SSL, and DASGA require an ambient space representation of data as input, the data samples are mapped to the Euclidean space $\mathbb{R}^D$ with the multidimensional scaling (MDS) algorithm \cite{cox2000multidimensional} using the normalized graph weight matrices. The dimension of the MDS embedding is empirically chosen as $D=5$. The target domain misclassification errors of the methods, which are averaged over $1000$ random trials, are presented in Figure \ref{fig:Facebook}. The balanced error rates are reported in these results in order to remove any bias due to the unequal presence of the two classes in the data set.
 
 Gender prediction on social network graphs is a challenging problem; nevertheless, the gender information seems to be implicitly encoded to some extent in the graphs of the two communities, which can also be observed by inspecting the variation of the label functions on the two graphs in Figure \ref{fig:Facebook_dataset}. Figure  \ref{fig:Facebook} suggests that the proposed GrALP method performs reasonably well in this challenging setup. In Figure \ref{fig:errors_facebook_target}, GrALP performs better than the other methods when the ratio of available target labels is very few; however, gets outperformed by SA and JGSA when more target labels are known. In domain adaptation applications, typically no or few target labels are available, and the proposed method seems to effectively exploit the information in the source domain under such conditions. In Figures \ref{fig:errors_facebook_source}-\ref{fig:errors_facebook_all_labeled}, GrALP mostly outperforms the other methods.  Most of the compared methods have a correct classification rate fluctuating around $50\%$ in this binary classification setting, indicating that they cannot extract any information at all from labeled data samples. The comparison of the results in Figures \ref{fig:errors_facebook_match} and \ref{fig:errors_facebook_part_labeled} is particularly interesting. The classification accuracy of GrALP does not improve much with the increase in the number of matches in Figure \ref{fig:errors_facebook_match} where the matched nodes are unlabeled, in contrast to its tendency to improve in Figure \ref{fig:errors_facebook_part_labeled} where the matches are partially labeled. The knowledge of the labels at the matched nodes is seen to be helpful in this data set, where the label function (gender) has a much faster variation on the graphs compared to the previous data sets. The general increase in the error rates of the methods at high match ratios in Figure \ref{fig:errors_facebook_all_labeled} is a rather unexpected behavior, which probably occurs due to some bias caused by the particular distribution of the matched nodes on the two graphs, which cannot be eliminated with random repetitions as the identities of the matched nodes (common users between the two communities) are fixed in this particular data set.
	
	



\subsection{Parameter Analysis of the Proposed Algorithm} 
We now analyze the sensitivity of the proposed GrALP algorithm to the choice of the algorithm parameters. First, we investigate the effect of the weight parameters $\mu$, ${\gamma}_s$, and ${\gamma}_t$ on the misclassification rate.  The Multilingual text data set is used in this experiment. 90\% of the source samples and 10\% of the target samples are labeled and 10\% of the graph nodes are matched. The target misclassification rates obtained with different combinations of $\mu$, ${\gamma}_s$, and ${\gamma}_t$ values are presented in Table \ref{tab:mu-gamma-effect}.  We observe that setting the parameters according to the rule of thumb $\mu = 10\gamma_s = 10\gamma_t$ yields good performance. This result is also confirmed on the MIT-CBCL and COIL-20 data sets. 

\begin{table}[]
\footnotesize
\centering
\caption{Effect of the $\mu$, $\gamma_s$ and $\gamma_t$ parameters on the misclassification rate in
the multilingual text data set. Misclassification rates are in percentage.}
\label{tab:mu-gamma-effect}
\begin{tabular}{|l|c|c|c|c|c|c|c|}
\hline
Parameters &$\mu=10^{-2}$&$\mu=10^{-1}$&$\mu=10^{0}$&$\mu=10^{1}$&$\mu=10^{2}$\\ \hline
$\gamma_s=\gamma_t=10^{-3}$&  \textbf{24.05} & 25.92 & 35.44 & 46.40 & 53.19 \\ \hline
$\gamma_s=\gamma_t=10^{-2}$& 30.36 & \textbf{24.59} & 26.17 & 36.11 & 45.39 \\ \hline
$\gamma_s=\gamma_t=10^{-1}$ & 70.37 & 31.31 & \textbf{23.73} & 26.25 & 33.47  \\ \hline
$ \gamma_s=\gamma_t=10^{0}$ & 83.33 & 73.92 & 33.55 & \textbf{25.88} & \textbf{24.73}  \\ \hline
$ \gamma_s=\gamma_t=10^{1}$ &83.33 & 83.33 & 77.60 & 55.76 & 33.84 \\ \hline
\end{tabular}
\end{table}

 \begin{figure}[th!]
		\centering
		\begin{subfigure}[b]{0.23\textwidth}
			\centering
		   \includegraphics[width=3.4cm]{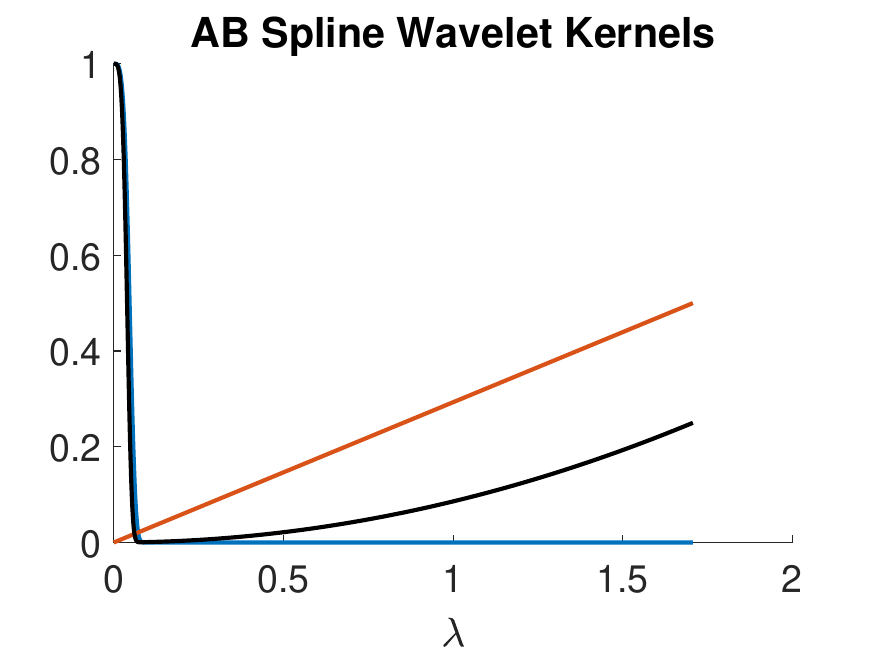}
		\caption{}
		\label{fig:WaveletKernels-abspline-1}
		\end{subfigure}%
~ 
		\begin{subfigure}[b]{0.23\textwidth}
			\centering
		   \includegraphics[width=3.4cm]{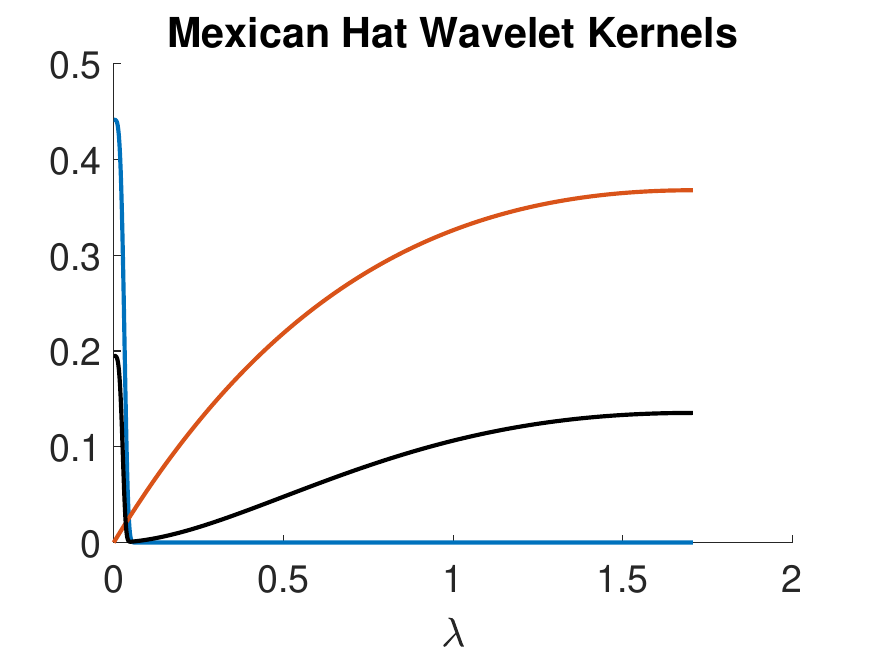}
		\caption{}
		\label{fig:WaveletKernels-mexican-1}
		\end{subfigure}%
~ 		
        \begin{subfigure}[b]{0.23\textwidth}
		\centering
		\includegraphics[width=3.4cm]{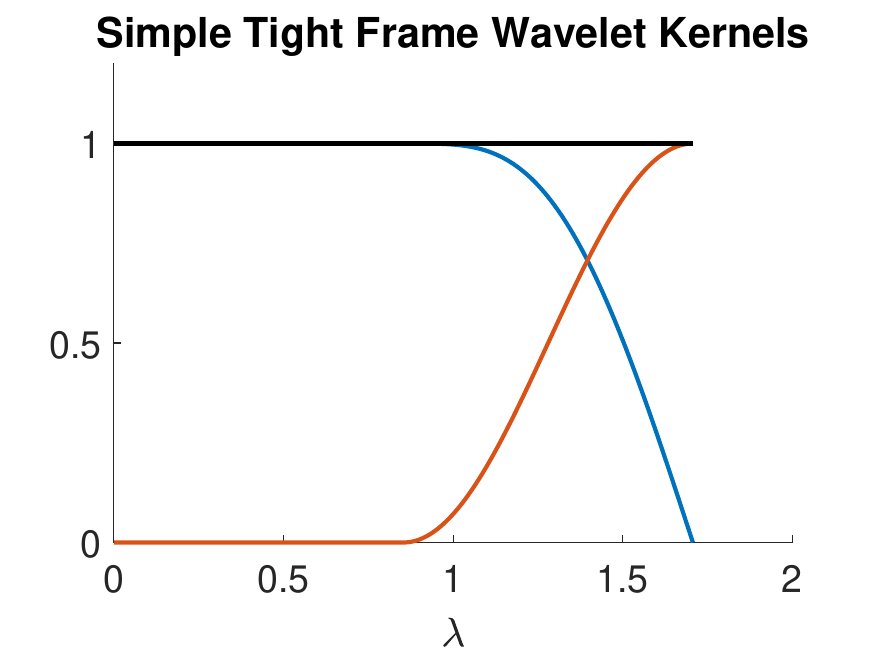}
		\caption{}
		\label{fig:WaveletKernels-simpletf-1}
		\end{subfigure}
~ 
        \begin{subfigure}[b]{0.23\textwidth}
			\centering
		   \includegraphics[width=3.4cm]{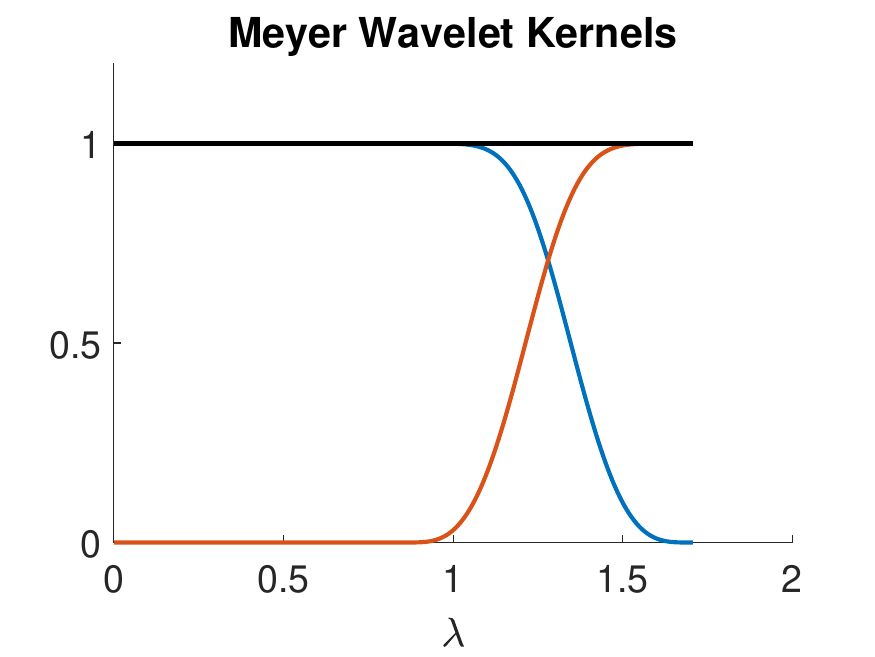}
		\caption{}
		\label{fig:WaveletKernels-meyer-1}
		\end{subfigure}%
        \\   
		\begin{subfigure}[b]{0.23\textwidth}
			\centering
			\includegraphics[width=3.4cm]{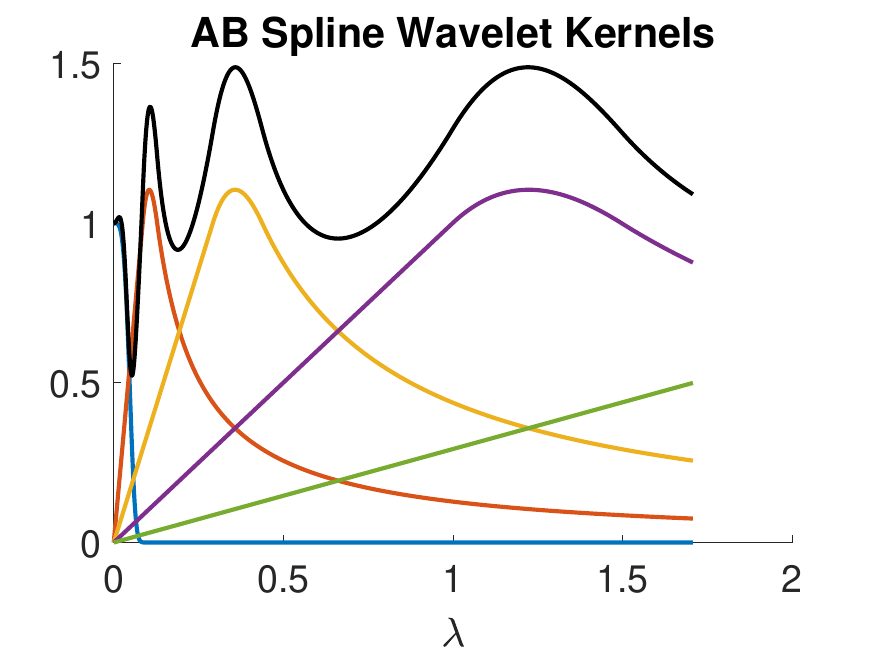}
			\caption{}
			\label{fig:WaveletKernels-abspline-4}
		\end{subfigure}	
				~ 
		\begin{subfigure}[b]{0.23\textwidth}
			\centering
			\includegraphics[width=3.4cm]{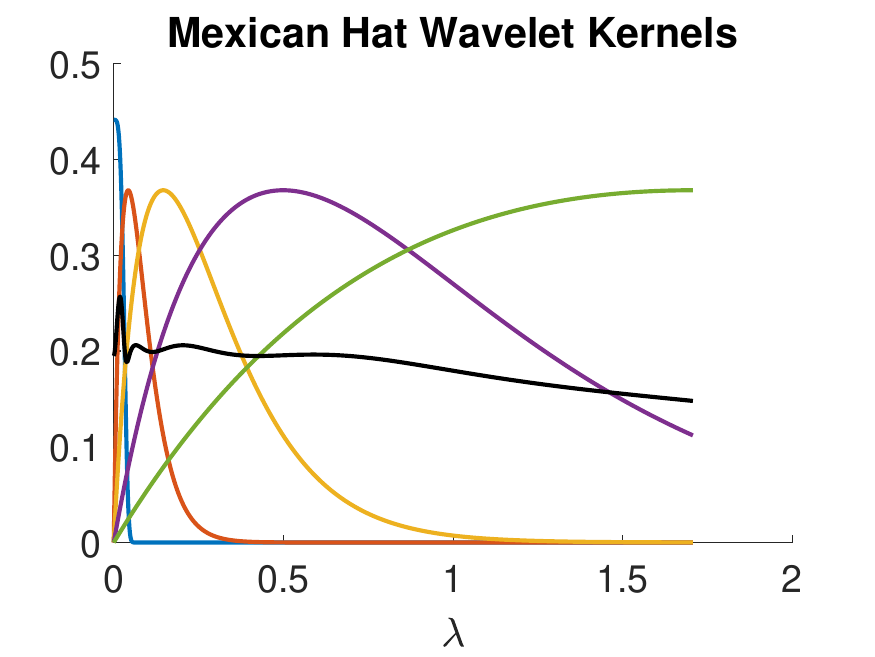}
			\caption{}
			\label{fig:WaveletKernels-mexican-4}
		\end{subfigure}		
				~ 
		\begin{subfigure}[b]{0.23\textwidth}
			\centering
			\includegraphics[width=3.4cm]{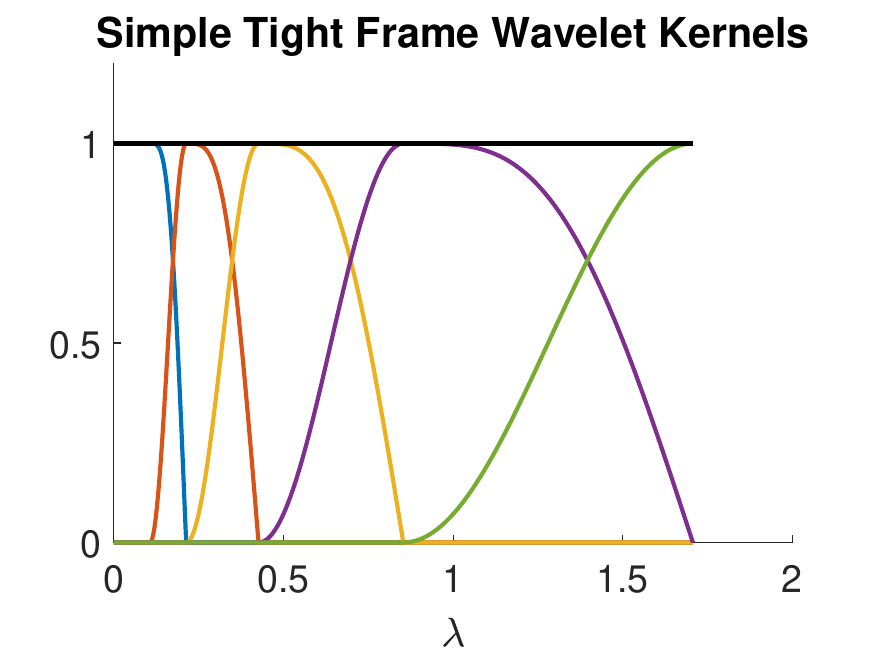}
			\caption{}
			\label{fig:WaveletKernels-simpletf-4}
		\end{subfigure}		
				~ 
		\begin{subfigure}[b]{0.23\textwidth}
			\centering
			\includegraphics[width=3.4cm]{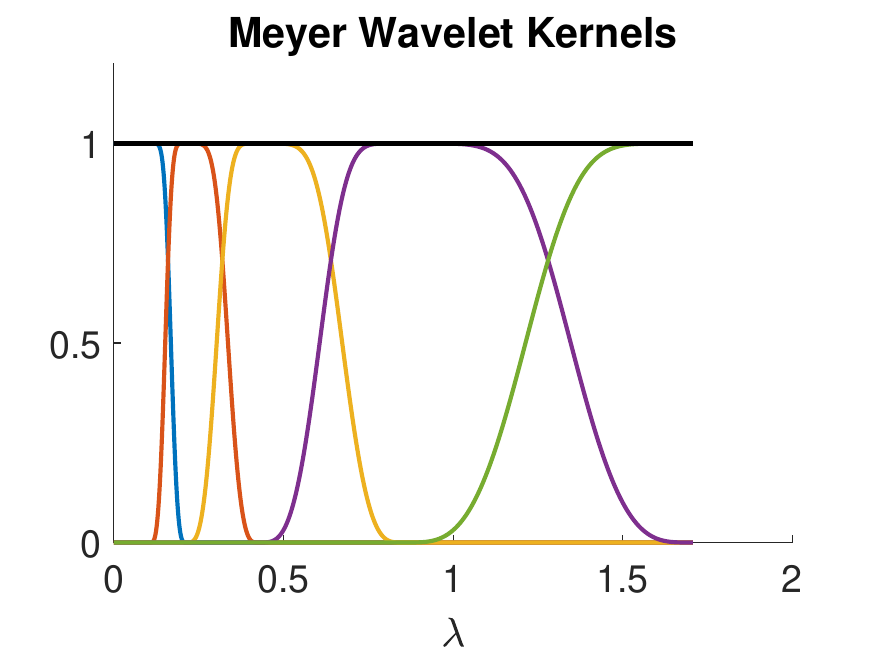}
			\caption{}
			\label{fig:WaveletKernels-meyer-4}
		\end{subfigure}   
		\caption{The spectral domain representations of the scaling and wavelet functions for AB spline, Mexican hat, Simple tight frame and Meyer kernels. The number of wavelets are chosen as 1 and 4, respectively in the upper and the lower rows. Black curves show the sums of the squares of the scaling and the wavelet functions.}
		\label{fig:WaveletKernels}
		\vspace{-0.4cm}
	\end{figure}

\begin{figure}[h]
		\centering
		\begin{subfigure}[t]{0.3\textwidth}
			\centering
			\includegraphics[height=3.0cm]{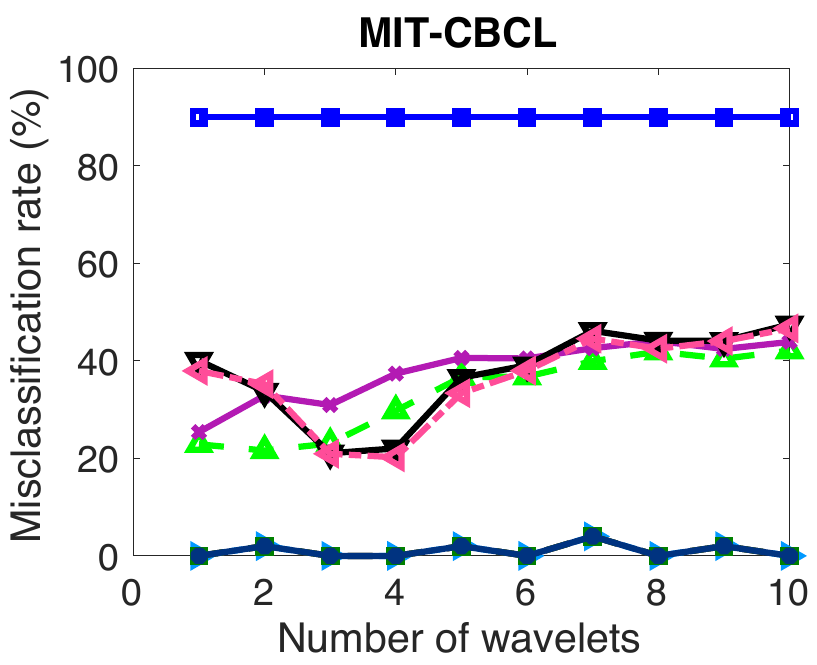}
			\caption{}
		\label{fig:WaveletAnalysis-MIT}
		\end{subfigure}
		~
		\begin{subfigure}[t]{0.3\textwidth}
			\centering
		   \includegraphics[height =3.0cm]{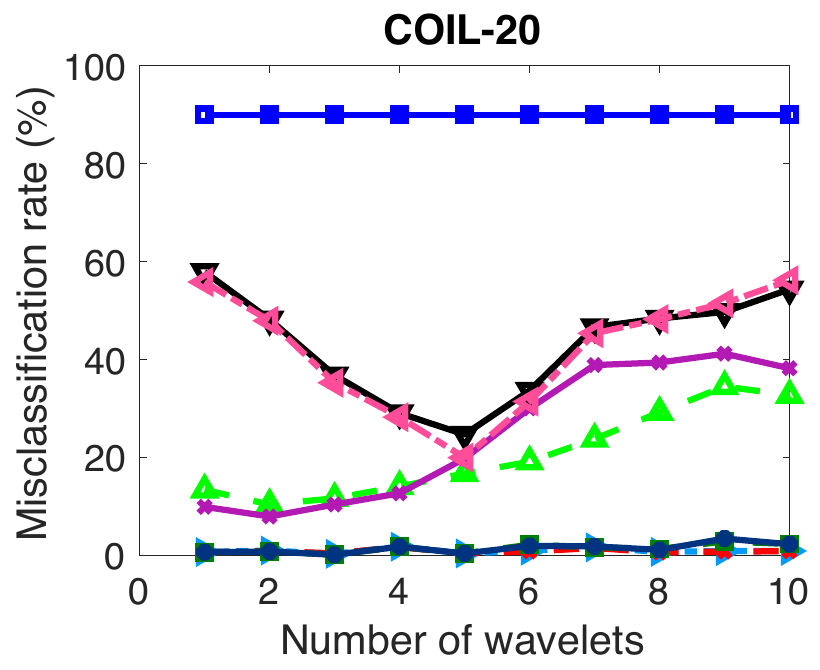}
		\caption{}
		\label{fig:WaveletAnalysis-COIL}
		\end{subfigure}%
		~
		\begin{subfigure}[t]{0.3\textwidth}
			\centering
			\includegraphics[height=3.0cm]{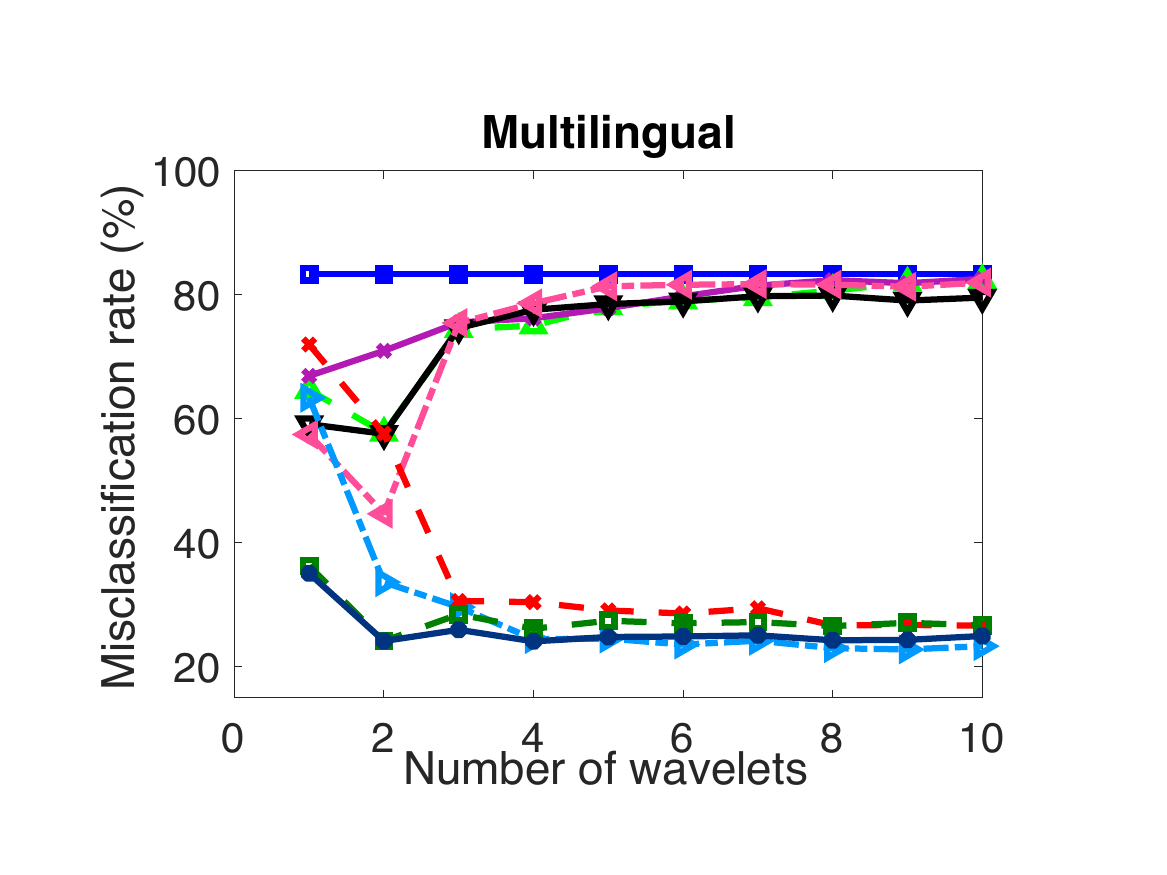}
			\caption{}
		\label{fig:WaveletAnalysis-Multilingual}
		\end{subfigure}\\
		~
		\begin{subfigure}[t]{0.3\textwidth}
			\centering
			\includegraphics[height=3.0cm]{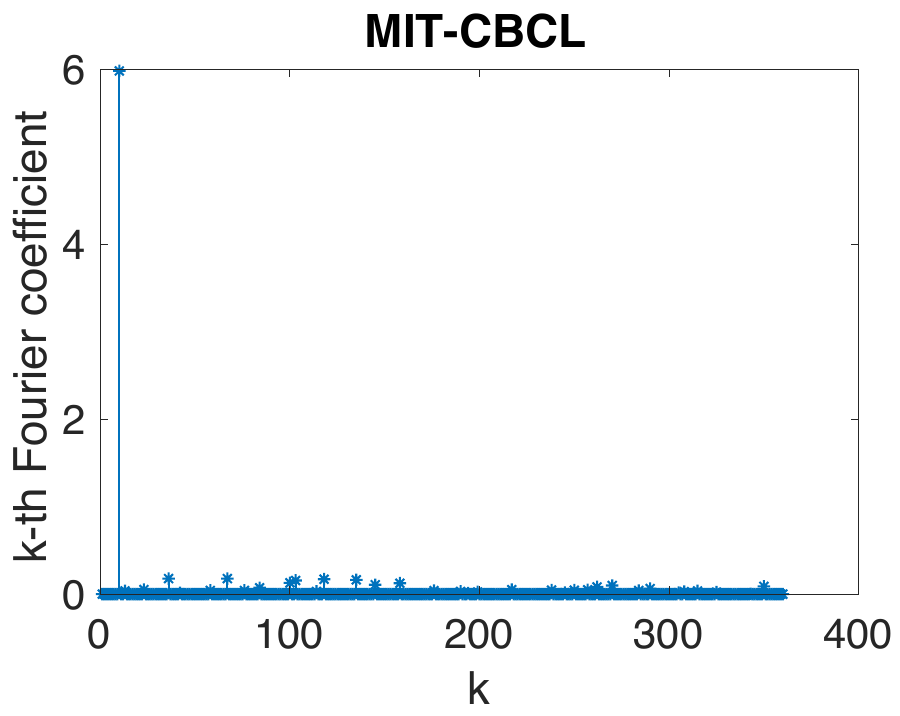}
			\caption{}
		\label{fig:Target_spec_mit}
		\end{subfigure}	
				~
		\begin{subfigure}[t]{0.3\textwidth}
			\centering
			\includegraphics[height=3.0cm]{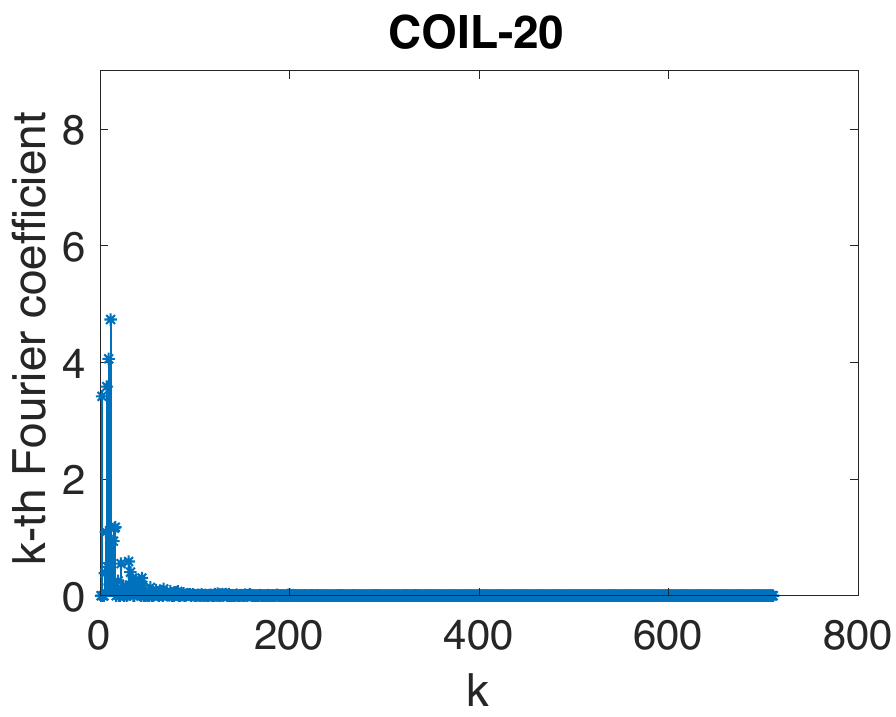}
		 \caption{}
		\label{fig:Target_spec_coil}
			\end{subfigure}	
					~
		\begin{subfigure}[t]{0.3\textwidth}
			\centering
			\includegraphics[height=3.0cm]{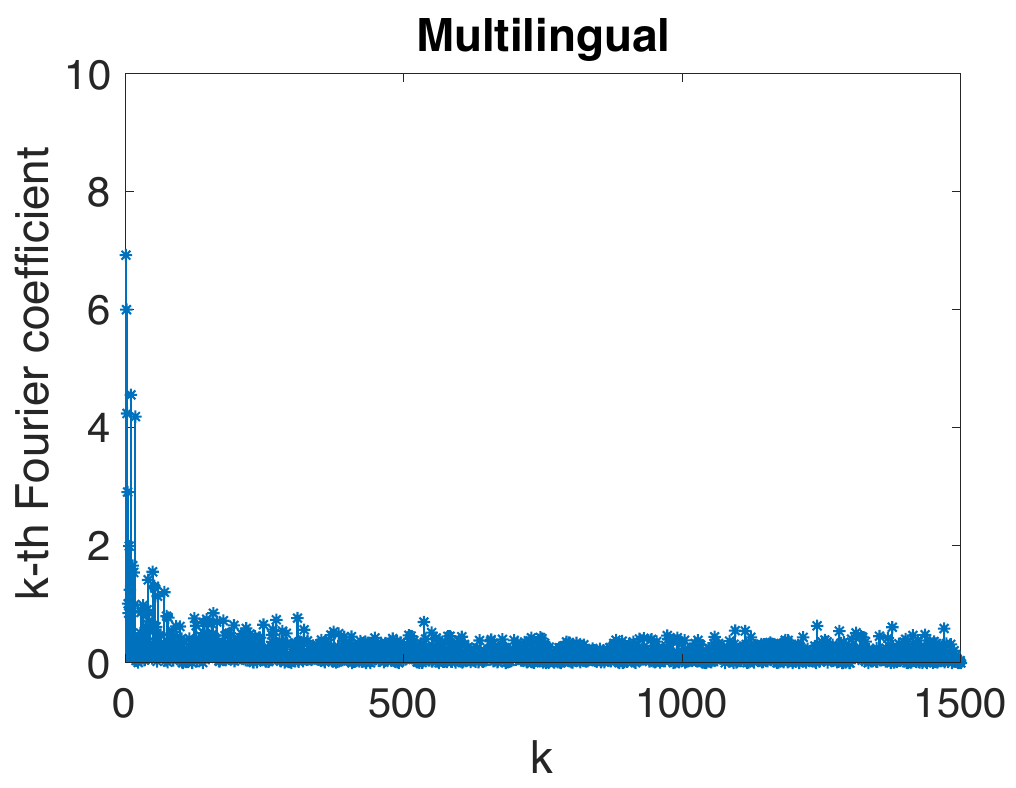}
		 \caption{}
		\label{fig:Target_spec_multiling}
		\end{subfigure}		
		~
		\begin{subfigure}[t]{0.3\textwidth}
			\centering
			\includegraphics[width=4cm]{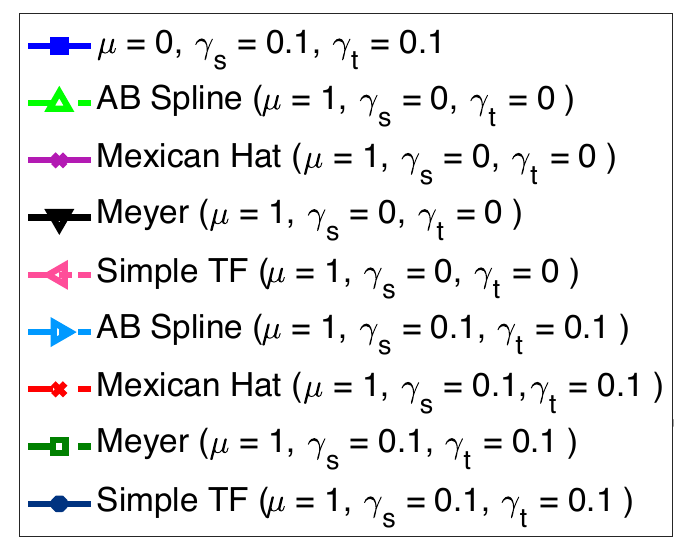}
		\end{subfigure}
		\caption{Effects of the wavelet kernel types and the number of wavelet functions on the misclassification rate.}
		\label{fig:WaveletAnalysis}
	\end{figure}

Next, we analyze how the misclassification rate is influenced by the choice of the wavelet kernel types and the number of wavelet functions used in the representation of the label functions. The four different wavelet kernel types (AB spline, Mexican hat, Simple tight frame, Meyer) provided by the Spectral Graph Wavelets Toolbox (SGWT) \cite{HammondVG11} are tested in our experiments. The scaling and wavelet functions given by these wavelet kernel types are shown in Figure \ref{fig:WaveletKernels} for different number of wavelets. 

The target misclassification rates obtained on the three data sets with these wavelet kernels are presented in Figure \ref{fig:WaveletAnalysis}. 90\%  of the source samples are labeled, 10\% of the nodes are matched, and no label information is available on the matched nodes or the target samples. Three different settings are tested with different combinations of the algorithm weight parameters. The magnitude of the Fourier coefficients of the label function is also plotted for each data set.

The results in Figures \ref{fig:WaveletAnalysis-MIT} and \ref{fig:WaveletAnalysis-COIL}  show that target labels can be predicted with very high accuracy in the MIT-CBCL and COIL-20 data sets for the choice of the weight parameters as $\mu=1$, $\gamma_s=0.1$, and  $\gamma_t=0.1$. The wavelet kernel types and the number of wavelet functions does not affect the misclassification rate much in this setting. The amount of information transferred from the source graph in addition to the smoothing effect of the regularization term is enough to obtain very high classification performance in these two data sets. In the Multilingual data set, the classification error decreases as the number of wavelets increases as seen in Figure \ref{fig:WaveletAnalysis-Multilingual}. The wavelet kernel type also affects the classification accuracy when the number of wavelets is small. The classification performance is better with the Simple tight frame or Meyer kernels, which are observed to have a better frequency coverage in Figure \ref{fig:WaveletKernels} compared to the AB Spline and Mexican hat kernels for a small number of wavelet functions. The AB spline and Mexican hat kernels have lower accuracy since these fail in representing certain parts of the spectrum, behaving like a band-stop filter if only one scaling function and one wavelet is used for instance. 

In the second setting, the regularization terms are removed by choosing $\gamma_s=\gamma_t=0$, in order to directly study the effect of the choice of the kernel type on the misclassification rate. An immediate observation in Figure \ref{fig:WaveletAnalysis} is that the performance degrades significantly in this second setting compared to the first one, which confirms the necessity of the regularization term. The results show that the misclassification rate first decreases and then increases in most data sets and kernel types in this setting. Using a too large number of wavelets degrades the performance for the following reason. As the number of wavelets increases, more and more wavelets capturing high frequency components of the label functions are involved in their representation, while transferring the high frequency information without regularization has an adverse effect as the high frequency part of the spectrum typically contains undesirable components such as noise or domain-specific variations. Examining the results for different data sets, we observe that AB spline and Mexican hat kernels perform better for the MIT-CBCL and COIL-20 data sets for a small number of wavelets. The plots in Figures \ref{fig:Target_spec_mit} and \ref{fig:Target_spec_coil} show that the spectra of the label functions are mostly concentrated at low frequencies in these two data sets, indicating that the label functions have a relatively slow variation on the graphs. The AB Spline and Mexican hat wavelets are more favorable in this case, as they are mostly concentrated around the low-frequency region of the spectrum when the number of wavelets is small. 
On the other hand, for the Multilingual data set, Meyer and Simple tight frame wavelets achieve better classification accuracy for a small number of wavelets. Due to the rather challenging structure of this data set, the label function varies relatively faster on the graphs, and has significant high-frequency components as seen in Figure \ref{fig:Target_spec_multiling}. Consequently, the Meyer and Simple tight frame kernels perform better, as they cover the high-frequency part of the spectrum better than the AB Spline and the Mexican hat wavelets.

Finally, in the third setting the parameters are set $\mu=0$ and $\gamma_s=\gamma_t=0.1$, in order to provide a comparison between our method and the reference solution that uses only the regularization term to predict the labels. As expected, this setting acts like a random classifier in all data sets, since there is no label information in the target domain and no information is transferred from the source domain in these experiments.  A global conclusion of all these experimental results is that the effect of the wavelet kernel types and the number of wavelets can vary among different data sets. The kernels should be carefully selected, considering the task at hand, the properties of the label function, and possibly the graph topology.

\section{Conclusion}
\label{sec:concl}

We have presented a domain adaptation method for classification problems defined on graph domains. The proposed method is based on the idea of sharing and transferring the information of the local characteristics of the label function between a source graph and a target graph, by using the projection coefficients of the label function onto spectral graph wavelet functions. Unlike conventional domain adaptation approaches relying largely on representations in a feature space, the proposed algorithm has minimal dependence on the feature space properties of data and treats the problem in an abstract graph setting. This leads to a flexible data representation model turning out to be advantageous for various domain adaptation problems that may be challenging to treat in the original data space. A mild assumption of the method is the availability of a small set of matches between the two graphs. Some future directions are the extension of the method to the case of unavailable match information, and the optional integration of available data feature vectors into the learning. 

\section{Acknowledgment}
This work has been partly supported by the T\"UB\.ITAK 2232 research scholarship under grant 117C007.

\bibliographystyle{IEEEbib}
\bibliography{refs}

\begin{thebibliography}{10}

\bibitem{HuangSGBS06}
J.~Huang, A.~J. Smola, A.~Gretton, K.~M. Borgwardt, and B.~Sch{\"{o}}lkopf,
\newblock ``Correcting sample selection bias by unlabeled data,''
\newblock in {\em Proc. Adv. Neur. Inf. Proc. Sys.}, 2006, pp. 601--608.

\bibitem{SunCPY11}
Q.~Sun, R.~Chattopadhyay, S.~Panchanathan, and J.~Ye,
\newblock ``A two-stage weighting framework for multi-source domain
  adaptation,''
\newblock in {\em Proc. Adv. Neur. Inf. Proc. Sys.}, 2011, pp. 505--513.

\bibitem{Fernando2013}
B.~Fernando, A.~Habrard, M.~Sebban, and T.~Tuytelaars,
\newblock ``Unsupervised visual domain adaptation using subspace alignment,''
\newblock in {\em Proc. IEEE Inf. Conf. Comp. Vis.}, 2013, pp. 2960--2967.

\bibitem{GongSSG12}
B.~Gong, Y.~Shi, F.~Sha, and K.~Grauman,
\newblock ``Geodesic flow kernel for unsupervised domain adaptation,''
\newblock in {\em Proc. {IEEE} Conf. Comp. Vision Pattern Rec.}, 2012, pp.
  2066--2073.

\bibitem{ZhangLO17}
J.~Zhang, W.~Li, and P.~Ogunbona,
\newblock ``Joint geometrical and statistical alignment for visual domain
  adaptation,''
\newblock in {\em Proc. {IEEE} Conf. Comp. Vis. Pat. Rec.}, 2017, pp.
  5150--5158.

\bibitem{PanTKY11}
S.~J. Pan, I.~W. Tsang, J.~T. Kwok, and Q.~Yang,
\newblock ``Domain adaptation via transfer component analysis,''
\newblock {\em {IEEE} Trans. Neural Networks}, vol. 22, no. 2, pp. 199--210,
  2011.

\bibitem{GhifaryBKZ17}
M.~Ghifary, D.~Balduzzi, W.~B. Kleijn, and M.~Zhang,
\newblock ``Scatter component analysis: {A} unified framework for domain
  adaptation and domain generalization,''
\newblock {\em {IEEE} Trans. Pattern Anal. Mach. Intell.}, vol. 39, no. 7, pp.
  1414--1430, 2017.

\bibitem{LongC0J15}
M.~Long, Y.~Cao, Y.~Wang, and M.~I. Jordan,
\newblock ``Learning transferable features with deep adaptation networks,''
\newblock in {\em Proc. 32nd Int. Conf. Mach. Learn.}, 2015, pp. 97--105.

\bibitem{WangD18}
M.~Wang and W.~Deng,
\newblock ``Deep visual domain adaptation: {A} survey,''
\newblock {\em Neurocomputing}, vol. 312, pp. 135--153, 2018.

\bibitem{ShumanNFOV13}
D.~I. Shuman, S.~K. Narang, P.~Frossard, A.~Ortega, and P.~Vandergheynst,
\newblock ``The emerging field of signal processing on graphs: Extending
  high-dimensional data analysis to networks and other irregular domains,''
\newblock {\em {IEEE} Signal Process. Mag.}, vol. 30, no. 3, pp. 83--98, 2013.

\bibitem{BronsteinBLSV17}
M.~M. Bronstein, J.~Bruna, Y.~LeCun, A.~Szlam, and P.~Vandergheynst,
\newblock ``Geometric deep learning: Going beyond euclidean data,''
\newblock {\em {IEEE} Signal Process. Mag.}, vol. 34, no. 4, pp. 18--42, 2017.

\bibitem{DongTRF19}
X.~Dong, D.~Thanou, M.~Rabbat, and P.~Frossard,
\newblock ``Learning graphs from data: {A} signal representation perspective,''
\newblock {\em {IEEE} Signal Process. Mag.}, vol. 36, no. 3, pp. 44--63, 2019.

\bibitem{HammondVG11}
D.~K. Hammond, P.~Vandergheynst, and R.~Gribonval,
\newblock ``Wavelets on graphs via spectral graph theory,''
\newblock {\em Applied and Computational Harmonic Analysis}, vol. 30, no. 2,
  pp. 129 -- 150, 2011.

\bibitem{PanY10}
S.~J. Pan and Q.~Yang,
\newblock ``A survey on transfer learning,''
\newblock {\em {IEEE} Trans. Knowl. Data Eng.}, vol. 22, no. 10, pp.
  1345--1359, 2010.

\bibitem{DaumeKS10}
H.~Daum{\'{e}}~III, A.~Kumar, and A.~Saha,
\newblock ``Co-regularization based semi-supervised domain adaptation,''
\newblock in {\em Proc. Adv. Neural Inf. Procesing Systems 23}, 2010, pp.
  478--486.

\bibitem{DuanXT12}
L.~Duan, D.~Xu, and I.~W. Tsang,
\newblock ``Learning with augmented features for heterogeneous domain
  adaptation,''
\newblock in {\em Proc. 29th International Conference on Machine Learning},
  2012.

\bibitem{Daume2010}
H.~Daum{\'e}, III, A.~Kumar, and A.~Saha,
\newblock ``Frustratingly easy semi-supervised domain adaptation,''
\newblock in {\em Proc. 2010 Workshop on Domain Adaptation for Natural Language
  Processing}, 2010, pp. 53--59.

\bibitem{CrammerKW08}
K.~Crammer, M.~Kearns, and J.~Wortman,
\newblock ``Learning from multiple sources,''
\newblock {\em Journal of Machine Learning Research}, vol. 9, pp. 1757--1774,
  2008.

\bibitem{WuWZTXYH17}
Q.~Wu, H.~Wu, X.~Zhou, M.~Tan, Y.~Xu, Y.~Yan, and T.~Hao,
\newblock ``Online transfer learning with multiple homogeneous or heterogeneous
  sources,''
\newblock {\em {IEEE} Trans. Knowl. Data Eng.}, vol. 29, no. 7, pp. 1494--1507,
  2017.

\bibitem{PereiraT18}
L.~A.~M. Pereira and R.~da~Silva~Torres,
\newblock ``Semi-supervised transfer subspace for domain adaptation,''
\newblock {\em Pattern Recognition}, vol. 75, pp. 235 -- 249, 2018.

\bibitem{LiangRZT19}
J.~Liang, R.~He, Z.~Sun, and T.~Tan,
\newblock ``Exploring uncertainty in pseudo-label guided unsupervised domain
  adaptation,''
\newblock {\em Pattern Recognition}, vol. 96, pp. 106996, 2019.

\bibitem{LopezPazHS12}
D.~L{\'{o}}pez{-}Paz, J.~M. Hern{\'{a}}ndez{-}Lobato, and B.~Sch{\"{o}}lkopf,
\newblock ``Semi-supervised domain adaptation with non-parametric copulas,''
\newblock in {\em Proc. Adv. Neur. Inf. Proc. Sys.}, 2012, pp. 674--682.

\bibitem{SunFS16}
B.~Sun, J.~Feng, and K.~Saenko,
\newblock ``Return of frustratingly easy domain adaptation,''
\newblock in {\em Proc. Thirtieth {AAAI} Conference on Artificial
  Intelligence}, 2016, pp. 2058--2065.

\bibitem{XuPXWLMS17}
Y.~Xu, S.~J. Pan, H.~Xiong, Q.~Wu, R.~Luo, H.~Min, and H.~Song,
\newblock ``A unified framework for metric transfer learning,''
\newblock {\em {IEEE} Trans. Knowl. Data Eng.}, vol. 29, no. 6, pp. 1158--1171,
  2017.

\bibitem{HerathHP17}
S~Herath, M.~T. Harandi, and F.~Porikli,
\newblock ``Learning an invariant {H}ilbert space for domain adaptation,''
\newblock in {\em Proc. {IEEE} Conf. Comp. Vis. Pat. Rec.}, 2017, pp.
  3956--3965.

\bibitem{TaoHW14}
J.~Tao, W.~Hu, and S.~Wang,
\newblock ``Sparsity regularization label propagation for domain adaptation
  learning,''
\newblock {\em Neurocomputing}, vol. 139, pp. 202 -- 219, 2014.

\bibitem{YangMY18}
B.~Yang, A.~J. Ma, and P.~C. Yuen,
\newblock ``Learning domain-shared group-sparse representation for unsupervised
  domain adaptation,''
\newblock {\em Pattern Recognition}, vol. 81, pp. 615 -- 632, 2018.

\bibitem{YaoPNLM15}
T.~Yao, Y.~Pan, C.~Ngo, H.~Li, and T.~Mei,
\newblock ``Semi-supervised domain adaptation with subspace learning for visual
  recognition,''
\newblock in {\em {IEEE} Conf. Comp. Vis. Pat. Rec}, 2015, pp. 2142--2150.

\bibitem{LuSC0H18}
H.~Lu, C.~Shen, Z.~Cao, Y.~Xiao, and A.~van~den Hengel,
\newblock ``An embarrassingly simple approach to visual domain adaptation,''
\newblock {\em {IEEE} Trans. Image Processing}, vol. 27, no. 7, pp. 3403--3417,
  2018.

\bibitem{LongZ0J17}
M.~Long, H.~Zhu, J.~Wang, and M.~I. Jordan,
\newblock ``Deep transfer learning with joint adaptation networks,''
\newblock in {\em Proc. 34th Int. Conf. Machine Learning}, 2017, pp.
  2208--2217.

\bibitem{TzengHSD17}
E.~Tzeng, J.~Hoffman, K.~Saenko, and T.~Darrell,
\newblock ``Adversarial discriminative domain adaptation,''
\newblock in {\em {IEEE} Conference on Computer Vision and Pattern
  Recognition}, 2017, pp. 2962--2971.

\bibitem{ChengP14}
L.~Cheng and S.~J. Pan,
\newblock ``Semi-supervised domain adaptation on manifolds,''
\newblock {\em {IEEE} Trans. Neural Netw. Learning Syst.}, vol. 25, no. 12, pp.
  2240--2249, 2014.

\bibitem{XiaoG15}
M.~Xiao and Y.~Guo,
\newblock ``Feature space independent semi-supervised domain adaptation via
  kernel matching,''
\newblock {\em {IEEE} Trans. Pattern Anal. Mach. Intell.}, vol. 37, no. 1, pp.
  54--66, 2015.

\bibitem{EynardKBGB15}
D.~Eynard, A.~Kovnatsky, M.~M. Bronstein, K.~Glashoff, and A.~M. Bronstein,
\newblock ``Multimodal manifold analysis by simultaneous diagonalization of
  {L}aplacians,''
\newblock {\em {IEEE} Trans. Pattern Anal. Mach. Intell.}, vol. 37, no. 12, pp.
  2505--2517, 2015.

\bibitem{RodolaCBTC17}
E.~Rodol{\`{a}}, L.~Cosmo, M.~M. Bronstein, A.~Torsello, and D.~Cremers,
\newblock ``Partial functional correspondence,''
\newblock {\em Comput. Graph. Forum}, vol. 36, no. 1, pp. 222--236, 2017.

\bibitem{ThanouSF14}
D.~Thanou, D.~I. Shuman, and P.~Frossard,
\newblock ``Learning parametric dictionaries for signals on graphs,''
\newblock {\em {IEEE} Trans. Signal Processing}, vol. 62, no. 15, pp.
  3849--3862, 2014.

\bibitem{ThanouF18}
D.~Thanou and P.~Frossard,
\newblock ``Learning of robust spectral graph dictionaries for distributed
  processing,''
\newblock {\em {EURASIP} J. Adv. Sig. Proc.}, vol. 2018, pp. 67, 2018.

\bibitem{PilanciV20}
M.~Pilanci and E.~Vural,
\newblock ``Domain adaptation on graphs by learning aligned graph bases,''
\newblock {\em IEEE Transactions on Knowledge and Data Engineering}, 2020.

\bibitem{Chung97}
F.~R.~K. Chung,
\newblock {\em Spectral Graph Theory},
\newblock American Mathematical Society, 1997.

\bibitem{HeinAv05}
M.~Hein, J.~Audibert, and U.~von Luxburg,
\newblock ``From graphs to manifolds - weak and strong pointwise consistency of
  graph laplacians,''
\newblock Max-Planck-Gesellschaft, 2005, pp. 470--485.

\bibitem{VetterliK95}
M.~Vetterli and J.~Kova\v{c}evic,
\newblock {\em Wavelets and Subband Coding},
\newblock Prentice-Hall, Inc., Upper Saddle River, NJ, USA, 1995.

\bibitem{CoifmanM06}
R.~R. Coifman and M.~Maggioni,
\newblock ``Diffusion wavelets,''
\newblock {\em Applied and Computational Harmonic Analysis}, vol. 21, no. 1,
  pp. 53 -- 94, 2006,
\newblock Special Issue: Diffusion Maps and Wavelets.

\bibitem{NarangO12}
S.~K. Narang and A.~Ortega,
\newblock ``Perfect reconstruction two-channel wavelet filter banks for graph
  structured data,''
\newblock {\em {IEEE} Trans. Signal Processing}, vol. 60, no. 6, pp.
  2786--2799, 2012.

\bibitem{leskovec2012learning}
J.~Leskovec and J.~J. Mcauley,
\newblock ``Learning to discover social circles in ego networks,''
\newblock in {\em Advances in neural information processing systems}, 2012, pp.
  539--547.

\bibitem{DonnatZHL18}
C.~Donnat, M.~Zitnik, D.~Hallac, and J.~Leskovec,
\newblock ``Learning structural node embeddings via diffusion wavelets,''
\newblock in {\em Proc. {ACM} {SIGKDD} Int. Conf. Knowledge Discovery {\&} Data
  Mining}, 2018, pp. 1320--1329.

\bibitem{MITCBCL}
B.~Weyrauch, B.~Heisele, H.~Huang, and V.~Blanz,
\newblock ``Component-based face recognition with {3D} morphable models,''
\newblock in {\em {IEEE} Computer Vision Pattern Rec. Workshop}, 2004, p.~85.

\bibitem{COIL-20}
S.~A. Nene, S.~K. Nayar, and H.~Murase,
\newblock ``Columbia object image library ({COIL}-20),''
\newblock Tech. {R}ep. CUCS-005-96, Department of Computer Science, Columbia
  University, February 1996.

\bibitem{Amini}
M.~R. Amini, N.~Usunier, and C.~Goutte,
\newblock ``Learning from multiple partially observed views -an application to
  multilingual text categorization,''
\newblock in {\em Proc. 22nd Int. Conf. Neural Information Processing Systems},
  2009, NIPS'09, pp. 28--36.

\bibitem{Ramos_usingtf-idf}
J.~Ramos,
\newblock ``Using {TF-IDF} to determine word relevance in document queries,''
\newblock in {\em Proc. 1st Instructional Conference on Machine Learning},
  2003.

\bibitem{ZhuGL03}
X.~Zhu, Z.~Ghahramani, and J.~D. Lafferty,
\newblock ``Semi-supervised learning using {G}aussian fields and harmonic
  functions,''
\newblock in {\em Proc. 20th Int. Conf. Machine Learning}, 2003, pp. 912--919.

\bibitem{cox2000multidimensional}
T.~F. Cox and M.~A. Cox,
\newblock {\em Multidimensional scaling},
\newblock Chapman and hall/CRC, 2000.

\end{thebibliography}

\end{document}